\documentclass{article}

\PassOptionsToPackage{numbers, compress}{natbib}

\usepackage[preprint]{neurips_2023}

\usepackage[utf8]{inputenc} 
\usepackage[T1]{fontenc}    
\usepackage{hyperref}       
\usepackage{url}            
\usepackage{booktabs}       
\usepackage{amsfonts}       
\usepackage{nicefrac}       
\usepackage{microtype}      
\usepackage{xcolor}         
\usepackage{amsmath}
\usepackage{graphicx}
\usepackage{amsfonts}
\usepackage{tipa}
\usepackage{verbatim}
\usepackage{subfigure}
\usepackage{booktabs,makecell,multirow}

\usepackage{float}
\usepackage{color}
\usepackage{bbm}
\usepackage{algorithm}
\usepackage{algorithmic}
\usepackage{multicol}
\usepackage{wrapfig}
\usepackage{bm}
\usepackage{array}
\usepackage{authblk}
\usepackage{setspace}
\usepackage{enumitem}
\usepackage{soul}
\usepackage{minitoc}

\newcommand{\ourmodel}{TCD}
\newcommand{\PCDM}{RTG-TS}
\newcommand{\HCDM}{HC-RTG-TS}
\newcommand{\ICDM}{FAR-RTG-TS}
\newcommand{\TFBB}{TFD}
\newcommand{\StateRewardDiffuser}{SRD}
\newcommand{\DistributionalQDiffuser}{DQD}
\newcommand{\RRDM}{RR-TCD}
\newcommand{\RQRDM}{RQR-TCD}

\title{Instructed Diffuser with Temporal Condition Guidance for Offline Reinforcement Learning}

\author{%
  $\text{\textbf{Jifeng Hu}}^{\bm{1}}$~~~~~~~~~~$\text{\textbf{Yanchao Sun}}^{\bm{2}}$~~~~~~~~~~$\text{\textbf{Sili Huang}}^{\bm{3}}$~~~~~~~~~~$\text{\textbf{SiYuan Guo}}^{\bm{4}}$~~~~~~~~~~$\text{\textbf{Hechang Chen}}^{\bm{5}}$\\
  $\text{\textbf{Li Shen}}^{\bm{6}}$~~~~~~~~~~$\text{\textbf{Lichao Sun}}^{\bm{7}}$~~~~~~~~~~$\text{\textbf{Yi Chang}}^{\bm{8}}$~~~~~~~~~~$\text{\textbf{Dacheng Tao}}^{\bm{9}}$\\
  ${}^{\bm{1,3,4,5,8}}$Jlilin University, Changchun, China\\
  ${}^{\bm{2}}$University of Maryland, College Park, MD 20742, USA\\
  ${}^{\bm{7}}$Lehigh University, Bethlehem, Pennsylvania, USA\\
  ${}^{\bm{6}}$JD Explore Academy, ${}^{\bm{9}}$The University of Sydney\\
  ${}^{\bm{1,3}}$\texttt{\{hujf21, huangsl21\}@mails.jlu.edu.cn}\\
  ${}^{\bm{2}}$\texttt{ycs@umd.edu}, ${}^{\bm{7}}$\texttt{lis221@lehigh.edu}\\
  ${}^{\bm{5,8}}$\texttt{\{chenhc, yichang\}@jlu.edu.cn}\\
  ${}^{\bm{4,6,9}}$\texttt{\{guosyjlu, mathshenli, dacheng.tao\}@gmail.com}\\
}

\begin{document}

\maketitle

\begin{abstract}
  Recent works have shown the potential of diffusion models in computer vision and natural language processing. 
  Apart from the classical supervised learning fields, diffusion models have also shown strong competitiveness in reinforcement learning (RL) by formulating decision-making as sequential generation.
  However, incorporating temporal information of sequential data and utilizing it to guide diffusion models to perform better generation is still an open challenge.
  In this paper, we take one step forward to investigate controllable generation with temporal conditions that are refined from temporal information. 
  We observe the importance of temporal conditions in sequential generation in sufficient explorative scenarios and provide a comprehensive discussion and comparison of different temporal conditions.
  Based on the observations, we propose an effective temporally-conditional diffusion model coined Temporally-Composable Diffuser (\ourmodel{}), which extracts temporal information from interaction sequences and explicitly guides generation with temporal conditions.
  Specifically, we separate the sequences into three parts according to time expansion and identify historical, immediate, and prospective conditions accordingly.
  Each condition preserves non-overlapping temporal information of sequences, enabling more controllable generation when we jointly use them to guide the diffuser.
  Finally, we conduct extensive experiments and analysis to reveal the favorable applicability of \ourmodel{} in offline RL tasks, where our method reaches or matches the best performance compared with prior SOTA baselines.
\end{abstract}

\section{Introduction}

Diffusion probabilistic models (DPMs) have shown impressive results in photo-realistic image synthesization~\cite{ho2020denoising, song2019generative, nichol2021improved}, text-to-image generation~\cite{kim2022diffusionclip, ramesh2022hierarchical}, and realistic video creation~\cite{esser2023structure, khachatryan2023text2video, ceylan2023pix2video}.
Besides, DPMs are not limited to classical supervised learning tasks mentioned above. 
More broadly, diffusion-based RL methods have also shown huge potential in sequential decision-making problems~\cite{wang2022diffusion, janner2022planning, fu2020d4rl}, facilitating many successful attempts in RL~\cite{janner2022planning}.
For example, \citet{ajay2022conditional} propose Decision Diffuser (DD), which learns policies with the return-conditioned, constraint-conditioned, or skill-conditioned diffuser and achieves better performance in many offline RL tasks.

Given the initial states, prior studies usually adopt heuristic conditions to generate behaviors by either the action-participated or non-action-participated diffusion process~\cite{kumar2020conservative, walke2023don}.
The former generation strategy directly generates the state-action sequences, while the latter method first synthesizes state sequences and then generates actions with inverse dynamics or other models~\cite{janner2022planning, chi2023diffusion}.
However, regardless of which approach to be adopted, the condition of the diffusion model always plays a pivotal role in generating plausible sequences where inappropriate conditions will lead to sub-optimal policies~\cite{wang2022diffusion, ajay2022conditional, lu2023synthetic}.
Heuristic conditions adopted by previous studies could cause several undesirable consequences because they do not fully consider temporal information, which is critical for understanding the dynamics, dependencies, and consequences of decisions over time in sequential modeling problems.
Although some existing approaches~\cite{ajay2022conditional, janner2022planning} are conditioned on \emph{prospective} information, such as future returns, they usually neglect the \emph{immediate} behaviors and the \emph{historical} behaviors, which are important during long sequence generation, especially in partially observable and highly stochastic environments.
Since temporal dependencies are associated with the performance of diffusion models, a key question arises:

\begin{center}
   \emph{How can we further dig into the potential of DPMs by considering the\\ temporal properties of decision-making in RL?} 
\end{center}

In this paper, we aim to identify temporal information from experiences, systematically understand the effects of temporal dependencies, and explicitly incorporate temporal conditions into the diffusion and generation processes of diffusion models.
Specifically, we identify three distinct classes: historically-conditional, immediately-conditional, and prospectively-conditional sequence generation.
Any arbitrary combination of these three conditions can be integrated, which we refer to as interchangeably-conditional sequence generation.
Then, we provide a unified discussion of temporal conditions about their respective advantages, disadvantages, and connections to existing works, including potential implementation approaches and corresponding experimental results.
Inspired by the above discoveries, we propose a generic temporally-conditional diffusion model and observe that this Temporally-Composable Diffuser (\ourmodel{}), with the diffusion model as a sequential planner and different temporal conditions as guidance, can capture the sequential distribution information and generate conditional planning trajectories.
We adopt classifier-free training, where the interchangeably-temporal conditions are composed with samples together during reverse denoising process, and perform generation by considering the interactive history, statistical current action rewards, and remaining available returns.
Additionally, apart from the above-mentioned temporal condition types, we also draw inspiration from recent works~\cite{chen2021decision, bellemare2017distributional, koenker2001quantile, andrychowicz2017hindsight}, and incorporate them into our proposed \ourmodel{} method to extend the attention of historical sequence length, perform better estimation on out-of-distribution actions, obtain radical and conservative reward estimation, and capture more useful feature information. 
The sufficient experiments confirm that \ourmodel{} can perform better than other baselines in various offline RL tasks, coinciding with our motivations and findings.

In summary, our main contributions are four-fold: 
\begin{itemize}[noitemsep,leftmargin=*]
\setlist{nolistsep} 
    \item We rethink the temporal dependencies of context sequences in diffusion models and find that current diffusion-based models with heuristic conditions can not fully dig into the potential of diffusion models and lead to sub-optimal performance.
    \item We propose the Temporally-Composable Diffuser (\ourmodel{}), which can capture the temporal dependencies of sequences when performing sequential generation. Furthermore, we provide a comprehensive discussion of the effects of temporal conditions, which reveals potential improvements and helps new algorithms discovery.
    \item Inspired by the discussion of temporal conditions, we also consider incorporating other techniques, such as transformer-backbone, distributional RL, quantile regression, and experience replay, into our method and provide some new variants with temporal condition guidance in Appendix~\ref{More Discussion about Temporal Conditions}.
    \item Finally, we conduct extensive experiments and discussions in Section~\ref{Experiments} to investigate the applicability of temporal conditions. 
    The results show that our method can surpass or match the best performance compared with other baselines.
\end{itemize}

\section{Related Work}

\noindent\textbf{Offline RL.}
Offline RL aims to learn the optimal policy from previously-collected datasets without extra interaction~\cite{levine2020offline, kostrikov2021offline, kumar2020conservative, wu2019behavior, ghosh2022offline, an2021uncertainty,ross2011reduction, liu2019off, siegel2020keep, an2021uncertainty}.
Although offline RL technology makes it possible for avoiding expensive and risky data collection process, in practice, the distribution shift between the learned policy and the data-collected policy poses difficulties for improving performance~\cite{kumar2020conservative, fujimoto2021minimalist}. 
The overestimation of out-of-distribution (OOD) actions produces errors in policy evaluation, which results in poor performance~\cite{levine2020offline, rezaeifar2022offline}.
In order to solve this issue, recent works can be roughly divided into two categories. 
Model-free offline RL methods apply constraints on learned policy and value function to prevent inaccurate estimation of unseen actions or enhance the robustness of OOD actions by introducing uncertainty quantification~\cite{xie2021bellman, kostrikov2021offline, kumar2019stabilizing}.
Model-based RL approaches propose to learn the optimal policy through planning or RL algorithms based on synthetic experiences generated from the learned dynamics~\cite{kidambi2020morel, rigter2022rambo, rafailov2021offline}.

\begin{figure*}[t!]
 \vspace{-0.3cm}
 \begin{center}
 \includegraphics[angle=0,width=0.95\textwidth]{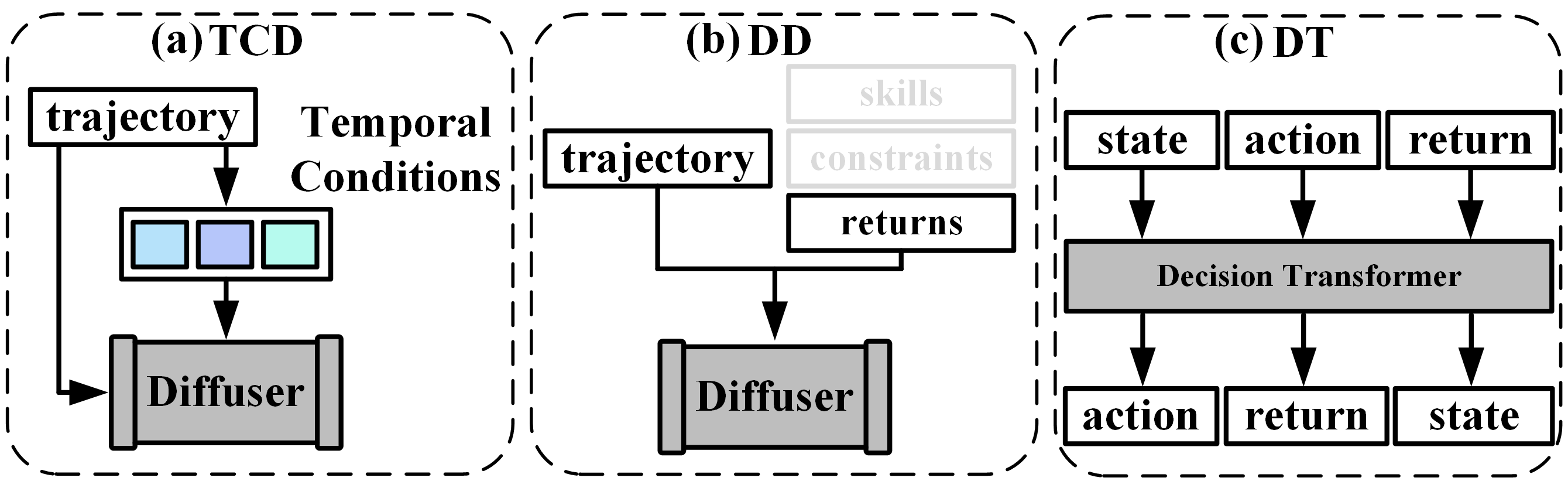}
 \caption{The comparison of \ourmodel{} and the other representative baselines, such as DD and DT.}
 \label{comparison with other baselines pipelines}
 \end{center}
 \vspace{-0.4cm}
 \end{figure*}

\noindent\textbf{Transformers in RL.}
Recent works show huge potential in RL by casting the decision-making as a sequence generation problem~\cite{chen2021decision, janner2021offline, zhang2023saformer, furuta2021generalized, wang2022bootstrapped,hu2022transforming}.
For example, given the future return as prompt, Decision Transformer (DT) orderly generates state, action, and reward tokens by considering the historical token sequence~\cite{chen2021decision}.
Another example is Trajectory Transformer (TT), which discretizes the state, action, and reward tokens and generates the sequences through beam search~\cite{janner2021offline}.
Compared with Transformer-based policies, the diffusion-based methods integrate planning and decision-making together, which leaves the value function estimation out.  Besides, the conditions of diffusion models guide the whole generative sequences directly, while the prompts of transformers work iteratively.

\noindent\textbf{Diffusion Probabilistic Models.}
Diffusion models have made big progress in image synthesis and text generation by formulating the data-generating process as an iterative denoising procedure~\cite{sohl2015deep, ho2020denoising, nichol2021improved, rombach2022high}. 
The denoising procedure can be derived from the posterior of the predefined diffusion process or the score matching of data distribution~\cite{song2019generative}.
In order to generate samples conditioned on human-preferred information, previous works propose to perturb the denoising process with classifier-guided methods and classifier-free methods~\cite{dhariwal2021diffusion, liu2023more, ho2022classifier}.
Though the classifier-guided method does not need retraining the diffusion model, More recent works reveal that the classifier-free guidance can generate better conditional samples~\cite{ajay2022conditional}.  
In this paper, we adopt classifier-free guidance, which is formed with an unconditional model and a conditional model, as the sampling method. 

\noindent\textbf{Diffusion Models in RL.}
Recently, offline decision-making problems have been formulated from the perspective of sequential distribution modeling and conditional generation with DPMs~\cite{janner2022planning, fontanesi2019reinforcement, ajay2022conditional, chi2023diffusion, chi2023diffusion, beeson2023balancing}, where high-performance policies are recovered by training on the given return-labeled trajectories datasets.  
This new pattern brings more flexible control in offline RL, such as goal-based planning, composable constraint combination, scalable trajectory generation, and complex skill synthesis~\cite{ajay2022conditional, wang2023pdpp}.
For example, \citet{janner2022planning} propose to combine the learned models and the trajectory optimization methods, effectively bypassing the adversarial examples that don't exist in the environment and reaching outstanding performance under proper condition guidance. 
\citet{ajay2022conditional} investigate how constraints and skills can be used to train DPMs and show the potential in many RL tasks. 
Additionally, diffusion policy is proposed as a more expressive policy, which has been used in RL, computer vision, and natural language processing~\cite{wang2022diffusion, chi2023diffusion, dai2023learning}.

Compared with existing works (Find Figure~\ref{comparison with other baselines pipelines} for synoptical comparison.), we are the first to consider temporal dependencies in generation with diffusion models.
We identify three types of temporal conditions, i.e., historical condition, immediate condition, and prospective condition, and propose a generic temporally-composable diffuser to produce behaviors with high performance.
Besides, we conduct extensive experiments and provide a comprehensive discussion of temporal conditions.

\section{Preliminaries}

In this section, we first present the relationship between decision-making and sequence generation.
Then we review the conditional generation with diffusion models. 

\subsection{Reinforcement Learning as Sequence Generation}

In classical reinforcement learning, the sequential decision-making problem is formulated via the Markov Decision Process (MDP), which is defined as the tuple $\mathcal{M}=\langle\mathcal{S}, \mathcal{A}, \mathcal{T}, \mathcal{R}, \rho_0, \gamma\rangle$ where $\mathcal{S}$ and $\mathcal{A}$ represent the state and action space, respectively, $\mathcal{T}: \mathcal{S}\times\mathcal{A}\rightarrow \Delta(\mathcal{S})$ denotes the Markovian transition probability, $\mathcal{R}: \mathcal{S}\times\mathcal{A}\times\mathcal{S}\rightarrow \mathbb{R}$ is the reward function, $\rho_0$ is the initial state distribution, and $\gamma\in [0, 1)$ is the discount factor.
At each time step $t$, the agent receives a state $s_t$ from the environment and produces an action $a_t$ with a stochastic or deterministic policy $\pi$. 
Then a reward $r_t=r(s_t, a_t)$ from the environment serves as the feedback to the executed action of the agent.
After the interactive interaction with the environment in a whole episode, we will obtain the state, action, and reward sequence $\tau=\{s_t,a_t,r_t\}_{t\geq 0}$.
In RL, our goal is to find a policy $\pi$ that can maximize the discounted return $\mathbb{E}_{\rho_0,\pi}[\sum_{t=0}^{\infty}\gamma^{t}r(s_t, a_t)]$~\cite{sun2022certifiably, sun2023smart}.

Each trajectory $\tau$ can be regarded as a data point sampled from trajectory distribution according to certain policy $\pi$.
Then we can use diffusion models to learn the data distribution $q(\tau)=\int q(\tau^{0:K}) d\tau^{1:K}$ with a predefined forward diffusion process $q(\tau^{k}|\tau^{k-1})=\mathcal{N}(\tau^{k};\sqrt{\alpha_k}\tau^{k-1},(1-\alpha_k)\bm{I})$ and a trainable generative process $p_{\theta}(\tau^{k-1}|\tau^{k})=\mathcal{N}(\tau^{k-1};\mu_{\theta}(\tau^{k},k), \Sigma_k)$, where $\alpha_{0:K}$ is preassigned, $\mu_{\theta}(\tau^{k})=\frac{1}{\sqrt{\alpha_k}}(\tau_k-\frac{\beta_k}{\sqrt{1-\bar{\alpha}_k}}\epsilon_{\theta}(\tau^{k},k))$, $\Sigma_k=\frac{1-\bar{\alpha}_{k-1}}{1-\bar{\alpha}_{k}}\beta_{k}\bm{I}$, and $\alpha_k + \beta_k = 1$.
Finally, the decision-making problem can be formulated as a sequential generation problem by learning a noising model $\epsilon_{\theta}(\tau^k, k)$ of the trajectory denoising process to capture the trajectory distribution and generate the offline datasets samples when a start state $s_t$ is given~\cite{sohl2015deep, ho2020denoising}.  
The simplified objective for training the diffusion model is defined by 
\begin{equation*}
    \mathcal{L}(\theta)=\mathbb{E}_{k\sim U(1, 2, ..., K), \epsilon\sim\mathcal{N}(0,\bm{I}), \tau^0\sim D}[||\epsilon-\epsilon_{\theta}(\tau^k, k)||_2^2],
\end{equation*}
where $k$ is the diffusion time step in the inverse diffusion process, $U$ denotes uniform distribution, $\epsilon$ denotes the multivariant Gaussian noise, $\tau^0=\tau$ is sampled from the replay buffer $D$, and $\theta$ is the parameters of model $\epsilon$. 

\subsection{Conditional Diffusion Probabilistic Models}\label{3.2}

There are two methods, classifier-guided and classifier-free, to train conditional diffusion models $p(\tau^{k-1}|\tau^{k}, \mathcal{C})$, i.e., generating data $\tau^{k-1}$ under perturbed variable $\tau^{k}$ and condition $\mathcal{C}$~\cite{dhariwal2021diffusion, liu2023more}.
The former method enables us to first train an unconditional diffusion model, which can be used to perform conditional generation under the gradient guidance of an additional classifier.
For the stochastic sampling process, such as DDPM~\cite{ho2020denoising}, $p_{\theta, \phi}(\tau^{k-1}|\tau^{k}, \mathcal{C})\propto p_{\theta}(\tau^{k-1}|\tau^{k})p_{\phi}(\mathcal{C}|\tau^{k})$ indicates that the classifier guidance information is $p_{\phi}(\mathcal{C}|\tau^{k})$, where the condition $\mathcal{C}$ should be the label of data $\tau^{k}$ and $\phi$ is the parameters.
Applying Taylor expansion on $log~p_{\phi}(\mathcal{C}|\tau^{k})$ at $\mu_{\theta}$, we can obtain the perturbed noise $\Sigma \cdot \nabla log~p_{\phi}(\mathcal{C}|\tau)$ that is added during the generation process.
Then we have $p(\tau^{k-1}|\tau^{k}, \mathcal{C})=\mathcal{N}(\mu_{\theta}+\Sigma_k \cdot \nabla log~p_{\phi}(\mathcal{C}|\tau), \Sigma_k)$.
For the deterministic sampling process, such as DDIM~\cite{song2020denoising}, the score function of joint distribution $p(\tau^{k},\mathcal{C})$ is defined by $\nabla_{\tau^{k}}log~(p_{\theta}(\tau^{k}, k)p_{\phi}(\mathcal{C}|\tau^{k}))=-\frac{1}{\sqrt{1-\bar{\alpha}_k}}\epsilon_{\theta}(\tau^k)+\nabla_{\tau^k}log~p_{\phi}(\mathcal{C}|\tau^{k})$.
The perturbed noise is $\epsilon_{\theta}(\tau^k, k)-\omega\sqrt{1-\bar{\alpha}_k}\nabla_{\tau^k}log~p_{\phi}(\mathcal{C}|\tau^{k})$, where $\omega$ is the guidance scale.

The classifier-free method builds the correlation between the samples and conditions in the training phase by learning unconditional noise $\epsilon_{\theta}(\tau^k, \emptyset, k)$ and conditional noise $\epsilon_{\theta}(\tau^k, \mathcal{C}, k)$, where we usually choose zero vector as $\emptyset$ in practice~\cite{ajay2022conditional}. 
Then the perturbed noise at each diffusion time step is calculated by $\epsilon_{\theta}(\tau^k, \emptyset, k)+\omega(\epsilon_{\theta}(\tau^k, \mathcal{C}, k)-\epsilon_{\theta}(\tau^k, \emptyset, k))$.
In this paper, we adopt the classifier-free guidance because it can usually bring more controllable generation and higher performance.

\begin{figure*}[t!]
 \begin{center}
 \includegraphics[angle=0,width=0.99\textwidth]{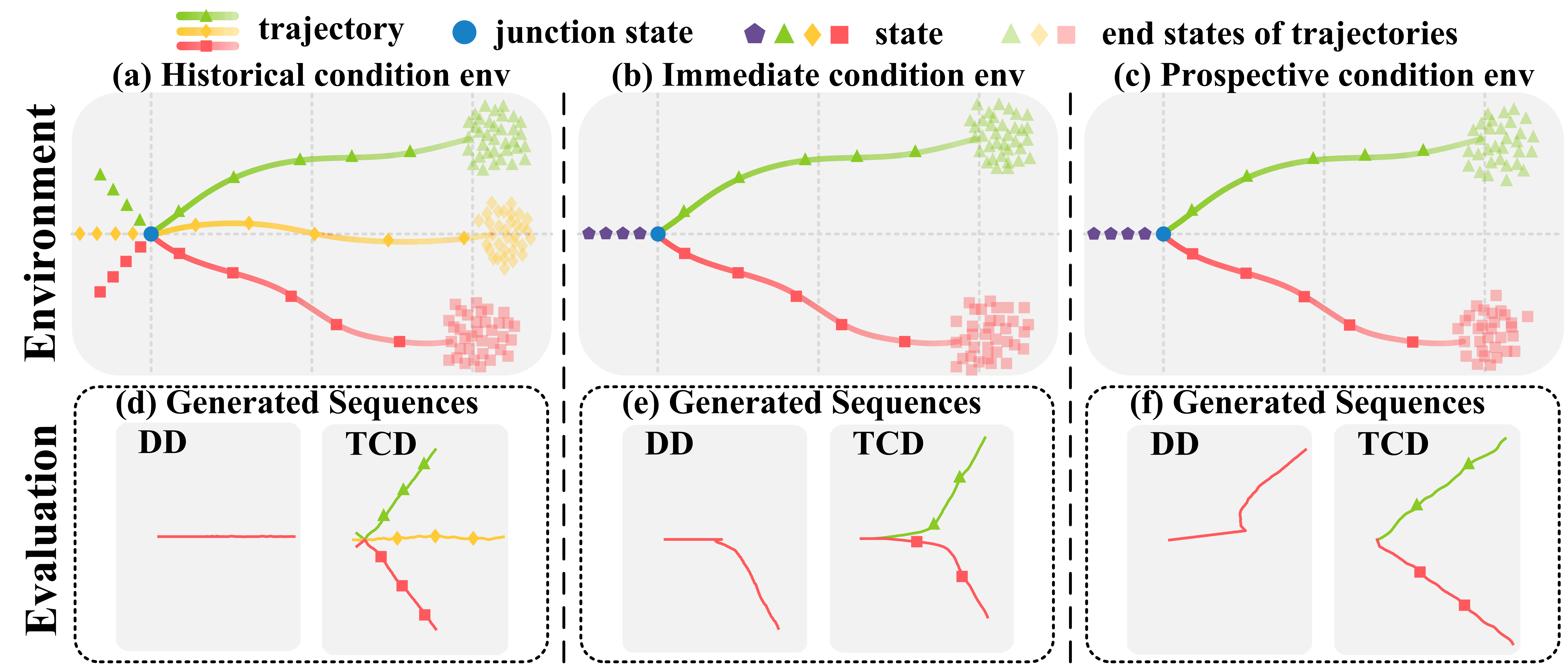}
 \caption{Temporal dependencies in RL. We design three types of RL environments (\textbf{a}, \textbf{b}, and \textbf{c}) to show the effect of temporal conditions in sequential modeling. The below figures (\textbf{d}, \textbf{e}, and \textbf{f}) are the corresponding evaluation, where the results show that \ourmodel{} can capture the temporal information, distinguish various sequences, and generate satisfactory sequences with temporal conditions.}
 \label{temporal condition env results}
 \end{center}
 \vspace{-0.4cm}
 \end{figure*}

\section{Rethink the Temporal Dependencies in Sequential Generation}\label{rethinking}
In this section, we rethink the temporal dependencies of sequences when regarding decision-making as the sequential generation with diffusion models and discuss the limitations of existing diffusers.

\textbf{What is the difference of generation between supervised learning and decision-making?}~~
Classical generation tasks, such as image synthesization, possess one-step property, where each picture does not have temporal dependencies.
But in decision-making tasks, each sequence contains temporal dependencies among the context transitions in it.
Roughly assimilating trajectory with image neglects the characteristics of multi-step interaction and temporal correlation of RL.
Besides, episodic returns are continuous values while the classes of figures are discrete values, which presents challenges for the generation when we only use one coarse-grained condition to guide the diffusion model.
Thus, in order to learn policies from the sequential data, we must extract temporal information and utilize it to guide the generation process of diffusion models.

\textbf{How temporal conditions enhance the performance in state sequence generation?}~~
To answer this question, we try to visualize the effects of temporal dependencies in sequential generation by conducting three types of environments (Figure~\ref{temporal condition env results}), which correspond to the temporal conditions.
In the historical condition env (Figure~\ref{temporal condition env results} (a)), the state sequences behind the junction state are strongly related to historical states, e.g., the green triangle trajectories only exist when the historical state sequence meets the condition of state incidence at the $135^{\circ}$ angle to the junction state.
In Figure~\ref{temporal condition env results} (b), we fix the historical sequences but just modify the action reward of different actions under the junction state, where the red square trajectories possess high rewards while low rewards arise in green triangle trajectories.
We fix the total reward the same in Figure~\ref{temporal condition env results} (c) by either masking the reward of the front part of trajectories with 0 or masking the latter part of trajectories with 0.
Refer to Appendix~\ref{Temporal Condition Scenarios} for the detailed environmental description.

The generation sequences of DD and our method (\ourmodel{}) under temporal conditions are shown in Figure~\ref{temporal condition env results} (d)-(f), where each below result is evaluated on the corresponding upper environment.
Taking Figure~\ref{temporal condition env results} (d) as an example, losing the capacity to capture temporal dependencies makes DD fail to generate homologous sequences based on historical states, while \ourmodel{} distinguishes the diverse samples and achieves controllable generation.
Although we may only need behaviors with high rewards in practice (Figure~\ref{temporal condition env results} (e)), the ability of subtle discernibility and diverse generations are as important as high performance.
Finally, in the results of Figure~\ref{temporal condition env results} (f), we can also see that the return-to-go (RTG) with remaining available time step can help diffusion model recognize different trajectories.
But the DD can not recover all types of trajectories when the prospective returns are the same.
More discussion can be found in Appendix~\ref{Discussion of Temporal Conditions Scenarios}.

\begin{figure*}[t!]
 \begin{center}
 \includegraphics[angle=0,width=0.99\textwidth]{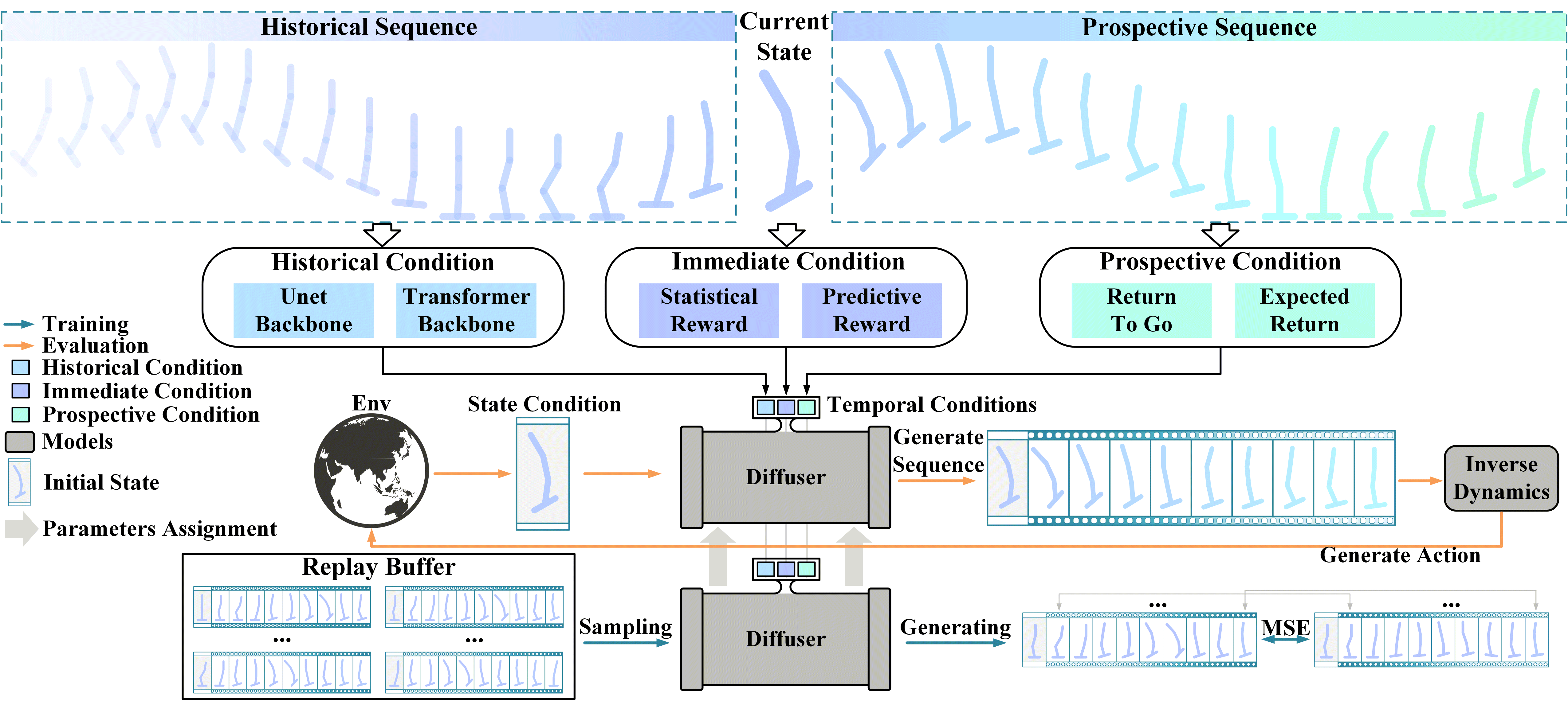}
 \caption{The overall architecture of \ourmodel{}. For the whole sequence shown on the top, we select one state as the current state. Then we can obtain the historical sequence and prospective sequence. During the training stage, we extract temporal information from the two separated sequences and the reward of the current state to train the diffusion model. During the evaluation stage, we adopt the temporal conditions that are obtained from the top-$Y$ trajectories to guide the generation process.}
 \label{architecture of our method}
 \end{center}
  \vspace{-0.4cm}
 \end{figure*}

\section{Temporally-Composable Diffuser}\label{our method}

In this section, we introduce the Temporally-Composable Diffuser (\ourmodel{}), as shown in Figure~\ref{architecture of our method}, which contains three types of temporal conditions from the perspective of temporal condition guidance ($\mathcal{C}_{TCG}$), and provide a unified discussion on these temporal conditions.
Due to the discrepancy across environments, various trajectory length makes it hard to train diffusion models, which need fixed sequence length. 
Following previous works~\cite{janner2022planning, ajay2022conditional}, the trajectories are split into equaling $\{s_t, a_t, r_t\}_{t:t+T-1}$ sequences (i.e., fragments of trajectory), where $T$ is the sequence length.
We use the a hat symbol to represent the generative sequence in the following parts.

During training, we use the diffusion model to capture the joint distribution of temporal condition guidance ($\mathcal{C}_{TCG}$) and the sequences. 
During generation, we search the temporal conditions from previously-collected experiences and interactive sequences. 
The universal training objective and generation process is given below
\begin{equation}\label{train DM}
    \mathcal{L}(\theta)=\mathbb{E}_{k\sim U(1, 2, ..., K), \epsilon\sim\mathcal{N}(0,\bm{I}), \tau^0\sim D}[||\epsilon-\epsilon_{\theta}(\tau_s^k,\mathcal{C}_{TCG}, k)||_2^2],
\end{equation}
\begin{equation}\label{eval DM}
    \tau_s^{k-1}=\frac{\sqrt{\bar{\alpha}_{k-1}}\beta_{k}}{1-\bar{\alpha}_k}\cdot\bar{\tau}_s+\frac{\sqrt{\alpha}_k (1-\bar{\alpha}_{k-1})}{1-\bar{\alpha}_{k}}\tau_s^{k}+|\Sigma_{k}|\bm{z},
\end{equation}
where $\mathcal{C}_{TCG}$ represents the mixture of prospective condition $\mathcal{C}_{PC}$, historical condition $\mathcal{C}_{HC}$, and immediate condition $\mathcal{C}_{IC}$.  $\bar{\epsilon}=\epsilon_{\theta}(\tau_s^k, \emptyset, k)+\omega(\epsilon_{\theta}(\tau_s^k, \mathcal{C}, k)-\epsilon_{\theta}(\tau_s^k, \emptyset, k))$, $\bar{\tau}_s=\frac{\tau_s^{k}-\sqrt{1-\bar{\alpha}_k}\bar{\epsilon}}{\sqrt{\bar{\alpha}_k}}$, $|\Sigma_k|=\frac{1-\bar{\alpha}_{k-1}}{1-\bar{\alpha}_{k}}\beta_{k}$, and $z\sim\mathcal{N}(\bm{0},\bm{I})$. Next, we introduce these temporal conditions successively.

\noindent\textbf{Prospective Condition $\mathcal{C}_{PC}$.}
For each sequence $\{\hat{s}_t, \hat{a}_t, \hat{r}_t\}_{t:t+T-1}$ to be waiting for generation, the prospective condition information can be provided as guidance, such as expected discounted return of state value or state-action value, RTG, and target goal state.
Compared with previous diffusion-based methods that estimate the action value function Q or state value function V, the advantage of RTG is that modeling one RTG bypasses inaccurate estimation on OOD samples.
Besides, RTG relates the remaining available time step with the historical best performance in one episode, which can not be reflected by Q or V function.

In this paper, we adopt the RTG, which indicates the future desired returns when preestablishing the initial returns.
In the generation process, we select the maximal return value in the datasets as the initial returns. 
Statistically, we first calculate the episode returns from the replay buffer and then obtain the maximal return value $\mathcal{T}_{max}$ and minimum return value $\mathcal{T}_{min}$.
After that, the RTG information from state $s_t$ is normalized by $\mathcal{J}_t=\frac{\mathcal{J}_t-\mathcal{T}_{min}}{\mathcal{T}_{max}-\mathcal{T}_{min}}$.
Finally, we use $[\mathcal{J}_t, t]$ as the condition during training, i.e., $\mathcal{C}_{PC}=[\mathcal{J}_t, t]$.
Furthermore, we find that our model can extrapolate to behavior sequences with higher returns when given a higher initial return offset. 
More discussion can be found in Section~\ref{Parameter Sensitivity Analysis}.
Similar to the processing way in Diffuser, where they apply the $s_t$ condition by replacing the denoised sequence $\{\hat{s}_t, \hat{a}_t, \hat{r}_t\}_{t:t+T-1}$ with $\{s_t, \hat{a}_t, \hat{r}_t, \{\hat{s}_{t+1}\hat{a}_{t+1}, \hat{r}_{t+1}\}_{t+1:t+T-1}\}$, we can also substitute the final generative state $\hat{s}_{t+T-1}$ with goal state $\hat{s}_{g}$, i.e., $\{s_t, \hat{a}_t, \hat{r}_t, \{\hat{s}_{t+1}\hat{a}_{t+1}, \hat{r}_{t+1}\}_{t+1:t+T-2}, s_g, \hat{a}_{t+T-1}, \hat{r}_{t+T-1}\}$.

\noindent\textbf{Historical Condition $\mathcal{C}_{HC}$.}
Classical diffusion model structure utilizes the U-net backbone and one-dimensional convolution to process sequence data, so the most straightforward method to consider historical information is conditioning on preceding experiences.
Specifically, based on the U-net backbone, we add the historical information by replacing the incipient generative state segments $\{\hat{s}_t, ..., \hat{s}_{t+T_{HC}-1}\}$ of sequences with $\mathcal{C}_{HC}=[\mathcal{H}]=\{s_t, ..., s_{t+T_{HC}-1}\}$ at each generative step.
Previous methods, like DD, generate state sequence conditioning only on current state $s_t$, which may omit important information that appears in history, especially for environments with partial observability.
Even for environments without partial observability, the historical conditions can be regarded as an argumentation method, and the experiments show they can still provide improvements.

\noindent\textbf{Immediate Condition $\mathcal{C}_{IC}$.}
Though we can use the diffusion model to plan a long-term sequence, only the first two states, $s_{t}$ and $\hat{s}_{t+1}$, are adopted to produce actions with inverse dynamics $a_t=f_{inv}(s_{t}, \hat{s}_{t+1})$.
Consequently, the direct influencing factor in obtaining rewards from the environment is the quality of the generated states $\hat{s}_{t+1}$.
This enlightens us that we should pay more attention to the current generative state $\hat{s}_{t+1}$, which is the meaning of immediate condition $\mathcal{C}_{IC}$.

The immediate condition $\mathcal{C}_{IC}$ works through two stages.
During the training stage, the first action reward $r_t$ and the corresponding sequence $\{s_t\}_{t: t+T-1}$ are bound together for training. 
Then in the evaluation stage, we select the $Y$ trajectories with top-$Y$ returns and extract the reward sequences $\tau_r$ as the immediate condition.
For each time step $t$ in evaluation, we use the $\tau_r[t]$ that associates $s_t$ and $a_t$ to instruct generation.
Thus, $\mathcal{C}_{IC}=[\tau_r[t]]$ is the immediate condition.

\section{Experiments}\label{Experiments}

In this section, we investigate the effects of different temporal conditions on a variety of different decision-making tasks~\cite{fu2020d4rl, fujimoto2019off, rajeswaran2017learning}.
We first introduce environmental settings in Section~\ref{Environment Settings}. Then in Section~\ref{Evaluation with baseliens} and Appendix~\ref{appendix Maze2D description}, we report the performance of \ourmodel{} on various tasks.
Next, we provide the discussion of temporal conditions in Section~\ref{Effects Analysis of Temporal Conditions} and Appendix~\ref{More Discussion about Temporal Conditions}.
Finally, we conduct parameter sensitivity analysis in Section~\ref{Parameter Sensitivity Analysis}.

 \begin{figure*}[t!]
 \begin{center}
 \includegraphics[angle=0,width=0.99\textwidth]{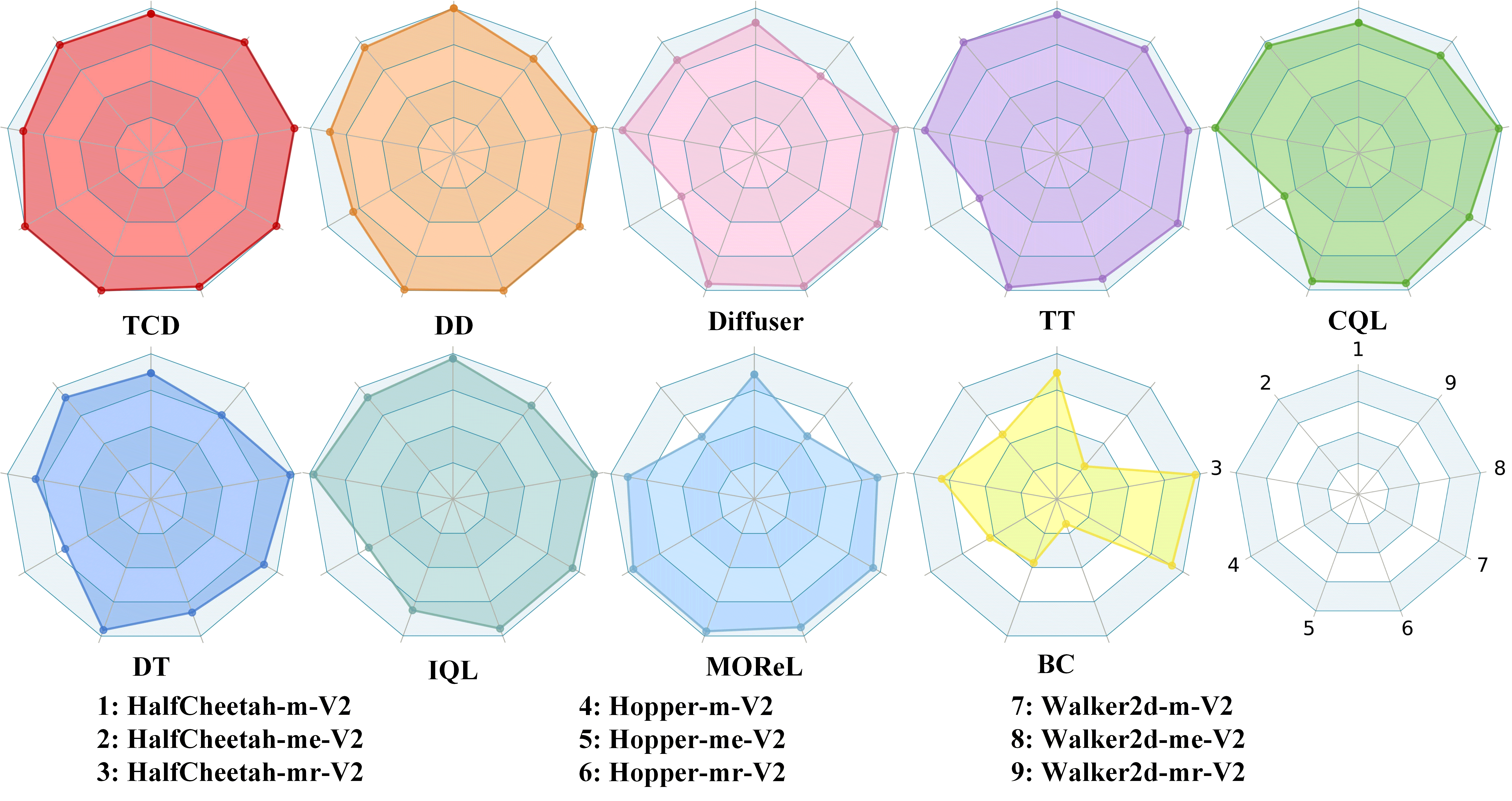}
 \caption{Offline RL algorithms comparison on D4RL Gym-MuJoCo datasets under successful comparison. The area of the polygon represents holistic performance. Each vertex of the polygon denotes the normalized performance across all models.}
 \label{comparison with baselines}
 \end{center}
 \vspace{-0.4cm}
 \end{figure*}

\subsection{Environment Settings} \label{Environment Settings}

\noindent\textbf{Environments.}\label{Environments} Gym-MuJoCo is a widely used benchmark of D4RL that provides offline datasets (HalfCheetah, Hopper, Walker2d) collected from single policy (-random (\textbf{-r}), -medium (\textbf{-m}), and -expert (\textbf{-e})) or multi-policies (-medium-replay (\textbf{-mr}), -medium-expert (\textbf{-me}), and -full-replay (\textbf{-fr}))~\cite{fu2020d4rl}.
The Maze2D is a 2D navigation task that requires the agent to reach the destination under sparse and dense reward functions~\cite{fujimoto2019off}.
See Appendix~\ref{appendix Maze2D description} for more description.

\noindent\textbf{Baselines.}\label{Baselines}
We compare our method with recent representative offline RL baselines (Figure~\ref{comparison with baselines}), including model-based methods MOReL~\cite{kidambi2020morel}, model-free methods CQL~\cite{kumar2020conservative}, IQL~\cite{kostrikov2021offline}, BC, and sequential modeling methods DT~\cite{chen2021decision}, TT~\cite{janner2021offline},  Diffuser~\cite{janner2022planning}, and DD~\cite{ajay2022conditional}.
In the analysis of temporal conditions, we also provide many other temporally-composable algorithms, such as diffuser with transformer-backbone (TFD), diffuser with state-reward sequence modeling (SRD), diffuser with distributional Q estimation (DQD), \ourmodel{} with reward estimation (RR-TCD and RQR-TCD) and other \ourmodel{} variants that remove certain temporal conditions.
Refer to Appendix~\ref{Additional Exploration about Temporal Conditions} for more description.

\noindent\textbf{Metrics.}\label{Metrics}
For all methods considered, we report the performance under different offline RL datasets. 
In order to compare the holistic capacity of all methods, the performance normalization based on all methods is considered because the value of performance in different scenarios varies considerably. 
For example, if we have evaluation results $(E_1, E_2, E_3)$ of three methods, then the normalized performance is defined by $E_i^{\prime}=\frac{\omega(E_i-min(E_1, E_2, E_3))}{max(E_1, E_2, E_3)-min(E_1, E_2, E_3)}$, where $i\in\{1, 2, 3\}$ and $\omega$ enables us to distinguish the differences of methods appropriately. In the experiments, we set $\omega=10$. 
Note that we also report the normalized score on environments in Appendix~\ref{additional experiments}.

\begin{figure*}[t!]
 \begin{center}
 \includegraphics[angle=0,width=0.99\textwidth]{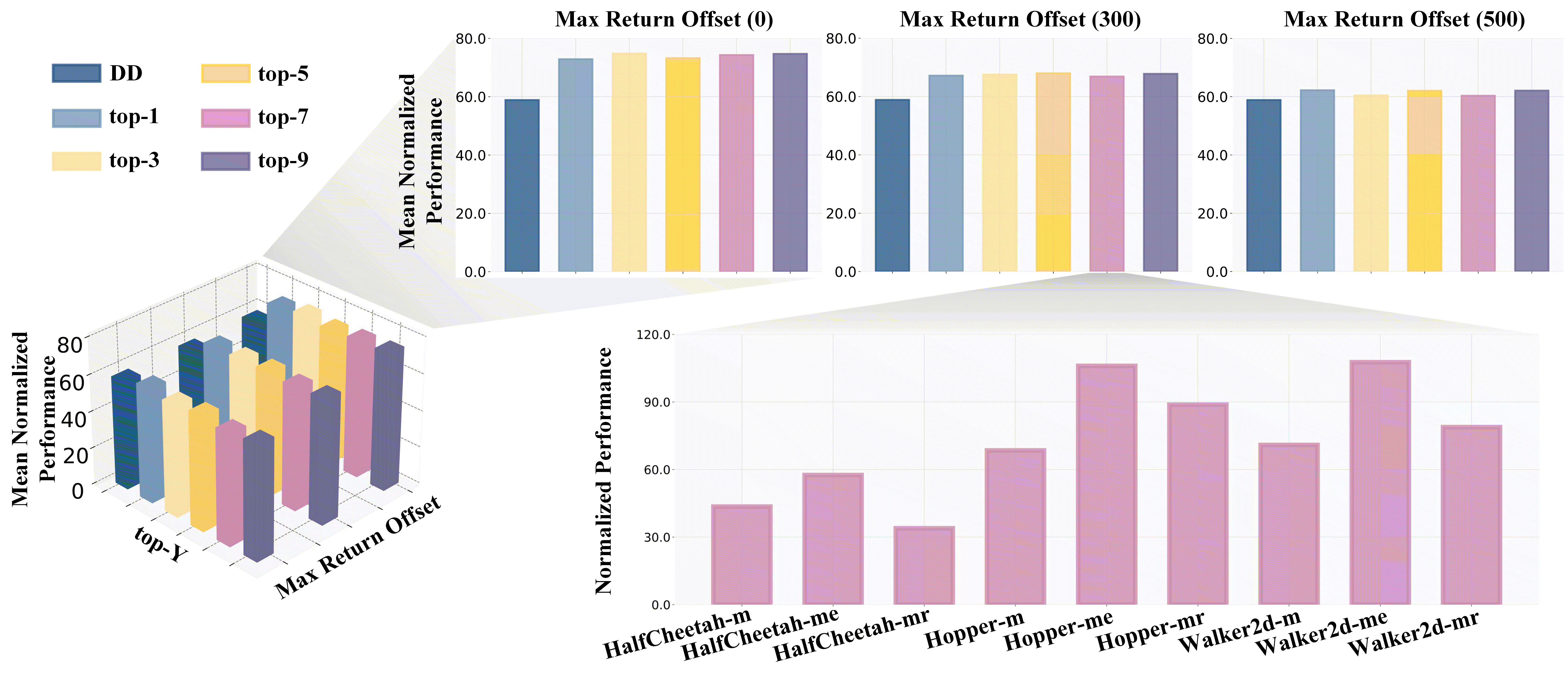}
 \caption{Parametric sensitivity of \ourmodel{} about top-$Y$ and max return offset under general comparison.}
 \label{sensitivity of top-K in general comparison}
 \end{center}
 \vspace{-0.4cm}
 \end{figure*}

\subsection{Evaluation on the Effectiveness of \ourmodel{}}\label{Evaluation with baseliens} 

\noindent\textbf{Evaluation on Temporal Condition Envs.}~~~~
We briefly introduce the environments in Section~\ref{rethinking}, report the experimental results in Figure~\ref{temporal condition env results}, and postpone more discussion in Appendix~\ref{Temporal Condition Scenarios} and \ref{Discussion of Temporal Conditions Scenarios}.

\noindent\textbf{Evaluation on Gym-MuJoCo.}~~~~
Results for the Gym-MuJoCo domains are shown in Figure~\ref{comparison with baselines} and Table~\ref{table comparison with baselines}.
The results for the baselines are based on the numbers reported in~\cite{ajay2022conditional}, \cite{janner2022planning}, and \cite{janner2021offline}.
On the datasets generated from single policy and multi policies, \ourmodel{} surpasses or matches the best prior methods in 8 of 9 environments under successful comparison, i.e., the evaluation time step is equal to the time limit of environments, and the agent is still capable of gaining more rewards.
Previous diffusion-based models (i.e., Diffuser and DD) do not realize the temporal dependencies among prospective, immediate, and historical decisions.
In comparison,  our method guides generation with these temporal conditions and outperforms baselines by large margins in Walker2d-mr and Hopper-m. 

\noindent\textbf{Evaluation on Maze2D.}~~~~In this experiment, we report the performance (Table~\ref{table comparison of maze2d}) of our method (\ourmodel{}) and baselines in the sparse and dense reward setting.
The sparse reward setting makes it hard for the agent to explore successful trajectories, especially in offline RL tasks, where the agent can not explore novel experiences through interaction with the environment.
As shown in Table~\ref{table comparison of maze2d}, in the Maze2D sparse setting, our method under temporal conditions guidance is consistently better than prior diffusion methods~\cite{ajay2022conditional}, such as DD, which generates sequences without temporal conditions guidance.
The reason is that the expected return estimation is almost zero for most states when the agent never or seldom observes successful trajectories, which results in restricted capacity in solving these sparse tasks.
In comparison, our method considers historical information, which provides wise decision reference when future instruction is relatively useful. 
With difficulty increasing from umaze to large under the Maze2D dense setting, the performance of DD degrades significantly, forming a sharp contrast that our method even reaches higher returns in larger size of environments.

\subsection{Ablation Study of Temporal Conditions}\label{Effects Analysis of Temporal Conditions} 
To investigate the effects of different temporal conditions, we report the performance of \ourmodel{} when removing certain temporal conditions.
Specifically, \PCDM{} denotes that we remove the historical condition and immediate condition during training and evaluation.
\HCDM{} denotes \ourmodel{} without immediate condition guidance, while \ICDM{} means that we cut the historical condition guidance.
We conduct the experiments on Gym-MuJoCo tasks under successful comparison.
As shown in Table~\ref{table ablation results under success}, each type of temporal condition contributes positive effects in the sequential generation.
Besides, the historical condition leads to greater performance gain compared with the immediate condition when considering the performance difference between \HCDM{} and \ICDM{}.
The reason can be attributed to the fact that historical condition provides multi-step information while immediate condition only contains one-step information.
More extensively, we consider other types of temporal conditions, including explicit and implicit.
The overall performance is shown in Figure~\ref{radar comparison on all temporal conditions}. 
The concrete values are summarized in Table~\ref{table comparison on all temporal conditions under general comparison}.
More discussion of the temporal conditions ranging from architecture backbone to training mode can be found in Appendix~\ref{More Discussion about Temporal Conditions}.

\subsection{Parameter Sensitivity Analysis}\label{Parameter Sensitivity Analysis} 
In this experiment, we investigate several hyper-parameters that may influence the performance of temporal conditions.
Specifically, we investigate the impacts of top-$Y$ and max return offset. 
Top-$Y$ denotes that the action-instructed reward sequence is selected from the trajectories of top-$Y$ expected return, while the max return offset represents the value that we add into the initial RTG setting during the evaluation stage. 
As shown in Figure~\ref{sensitivity of top-K in general comparison}, small $Y$, such as $Y=1$, is good enough for generating plausible sequences, but due to poor robustness to deflected situations, small $Y$ may result in low performance in Walker2d-m.
When we set a larger value of max return offset, the model may show over-optimistic to future return and weaken the effects of the current situation and historical behaviors, thus leading to performance decay.
But the above phenomenon also relates to the quality of datasets where relatively increasing the value of max return offset will bring higher performance.
For more discussion and experiments, please refer to Appendix~\ref{Additional Experiments about Parameter Sensitivity}.

\begin{table*}[t!]
\centering
\small
\caption{Offline RL algorithms comparison on Gym-MuJoCo datasets under successful comparison.}
\label{table comparison with baselines}
\resizebox{\textwidth}{!}{
\begin{tabular}{l | r r r | r r r | r r r | r}
\toprule
\specialrule{0em}{1.5pt}{1.5pt}
\toprule
Dataset & \multicolumn{3}{c|}{Med-Expert} & \multicolumn{3}{c|}{Medium} & \multicolumn{3}{c|}{Med-Replay} & \multirow{2}{*}{score}\\
\cline{1-10}
\rule{0pt}{2.5ex} Env & HalfCheetah & Hopper & Walker2d & HalfCheetah & Hopper & Walker2d & HalfCheetah & Hopper & Walker2d & \\
\midrule[1pt]
BC          & 55.2 & 52.5  & 107.5 & 42.6 & 52.9 & 75.3 & 36.6 & 18.1 & 26.0 & 51.9\\
MOReL       & 53.3 & 108.7 & 95.6  & 42.1 & 95.4 & 77.8 & 40.2 & 93.6 & 49.8 & 72.9\\
DT          & 86.8 & 107.6 & 108.1 & 42.6 & 67.6 & 74.0 & 36.6 & 82.7 & 66.6 & 74.7\\
Diffuser    & 79.8 & 107.2 & 108.4 & 44.2 & 58.5 & 79.7 & 42.2 & 96.8 & 61.2 & 75.3\\
IQL         & 86.7 & 91.5  & 109.6 & 47.4 & 66.3 & 78.3 & 44.2 & 94.7 & 73.9 & 77.0\\
CQL         & 91.6 & 105.4 & 108.8 & 44.0 & 58.5 & 72.5 & \textbf{45.5} & 95.0 & 77.2 & 77.6\\
TT          & 95   & 110.0 & 101.9 & 46.9 & 61.1 & 79.0 & 41.9 & 91.5 & 82.6 & 78.9\\
DD          & 90.6\tiny{$\pm$1.3} & 111.8\tiny{$\pm$1.8} & 108.8\tiny{$\pm$1.7} & \textbf{49.1}\tiny{$\pm$\textbf{1.0}} & 79.3\tiny{$\pm$3.6} & \textbf{82.5}\tiny{$\pm$\textbf{1.4}} & 39.3\tiny{$\pm$4.1} & \textbf{100}\tiny{$\pm$\textbf{0.7}} & 75\tiny{$\pm$4.3} & 81.8\\
\ourmodel{} & \textbf{92.67}\tiny{$\pm$\textbf{3.37}} & \textbf{112.60}\tiny{$\pm$\textbf{1.03}} & \textbf{111.31}\tiny{$\pm$\textbf{0.73}} & 47.20\tiny{$\pm$0.74} & \textbf{99.37}\tiny{$\pm$\textbf{0.60}} & 82.06\tiny{$\pm$1.83} & 40.57\tiny{$\pm$1.39} & 97.20\tiny{$\pm$2.39} & \textbf{88.04}\tiny{$\pm$\textbf{1.92}} & \textbf{85.67}\\
\bottomrule
\specialrule{0em}{1.5pt}{1.5pt}
\bottomrule
\end{tabular}}
\end{table*}

\begin{table*}[t!]
\centering
\small
\caption{Ablation study on temporal conditions under successful comparison.}
\label{table ablation results under success}
\resizebox{\textwidth}{!}{
\begin{tabular}{l | r r r | r r r | r r r | r}
\toprule
\specialrule{0em}{1.5pt}{1.5pt}
\toprule
Dataset & \multicolumn{3}{c|}{Med-Expert} & \multicolumn{3}{c|}{Medium} & \multicolumn{3}{c|}{Med-Replay} & \multirow{2}{*}{score}\\
\cline{1-10}
\rule{0pt}{2.5ex} Env & HalfCheetah & Hopper & Walker2d & HalfCheetah & Hopper & Walker2d & HalfCheetah & Hopper & Walker2d\\
\midrule[1pt]
\PCDM{} & 89.11\tiny{$\pm$4.47} & 113.13\tiny{$\pm$1.55} & 101.60\tiny{$\pm$9.90} & 43.58\tiny{$\pm$1.26} & 98.66\tiny{$\pm$1.03} & 79.23\tiny{$\pm$3.09} & 42.01\tiny{$\pm$1.23} & 97.27\tiny{$\pm$0.00} & 81.26\tiny{$\pm$4.30} & 82.87\\
\ICDM{} & 91.56\tiny{$\pm$2.80} & 113.63\tiny{$\pm$1.98} & 109.62\tiny{$\pm$0.66} & 44.48\tiny{$\pm$0.72} & 98.05\tiny{$\pm$2.00} & 78.55\tiny{$\pm$3.65} & 40.61\tiny{$\pm$1.58} & 99.21\tiny{$\pm$1.59} & 89.18\tiny{$\pm$2.65} & 84.98\\
\HCDM{} & 93.44\tiny{$\pm$1.62} & 112.51\tiny{$\pm$0.46} & 108.75\tiny{$\pm$0.31} & 44.63\tiny{$\pm$0.63} & 99.91\tiny{$\pm$0.35} & 83.16\tiny{$\pm$1.70} & 40.49\tiny{$\pm$1.42} & 98.34\tiny{$\pm$1.61} & 86.57\tiny{$\pm$2.11} & 85.31\\
\ourmodel{} & 92.67\tiny{$\pm$3.37} & 112.60\tiny{$\pm$1.03} & 111.31\tiny{$\pm$0.73} & 47.20\tiny{$\pm$0.74} & 99.37\tiny{$\pm$0.60} & 82.06\tiny{$\pm$1.83} & 40.57\tiny{$\pm$1.39} & 97.20\tiny{$\pm$2.39} & 88.04\tiny{$\pm$1.92} & 85.67\\
\specialrule{0em}{1.5pt}{1.5pt}
\bottomrule
\specialrule{0em}{1.5pt}{1.5pt}
\bottomrule
\end{tabular}}
\vspace{-0.4cm}
\end{table*}

\section{Conclusion, Limitations, and Broader Impact}
In this paper, we propose Temporally-Composable Diffuser (TCD), a generic temporally-conditional diffusion model that can extract temporal dependencies of sequences and achieve better controllable generation in sequential modeling.
We identify the historical conditions, immediate conditions, and prospective conditions from sequences and provide a comprehensive discussion and comparison of different temporal conditions.
We evaluate \ourmodel{} on extensive experiments, including several self-designed tasks and the D4RL datasets, where experiments demonstrate the superiority of our method compared with other sequential modeling methods and non-sequential modeling methods.

In terms of limitations, the mechanism of the generation process makes it slower than other models, such as Transformer-based models and MLP-based agents, even though we can use recent breakthroughs~\cite{nichol2021improved} to accelerate this process.
More recent studies have brought hope for the efficient generation. 
Thus we may be able to improve the efficiency based on the current models~\cite{song2023consistency}.
Refer to Appendix~\ref{More Discussion about Limitation and Future Work} for more discussions of limitations and future work.
Regarding societal impacts, we do not anticipate any negative consequences from using our method in practice.

\clearpage
\bibliography{main}
\bibliographystyle{plainnat}

\clearpage
\appendix

\clearpage
\section*{Appendix}

\section{Pseudocode of Temporally-Composable Diffuser (\ourmodel{})}\label{pseudocode}

The pseudocode for \ourmodel{} training is shown in Algorithm~\ref{ourmodel algorithm}.

\begin{algorithm}
    \caption{Temporally-Composable Diffuser (\ourmodel{})}
    \label{ourmodel algorithm}
    \begin{algorithmic}[1]
        \STATE \textbf{Input:} Diffusion model noise prediction model $\epsilon_{\theta}$, Inverse dynamics model $f_{\phi}$
        \STATE \textbf{Output:} $\epsilon_{\theta}$, $f_{\phi}$
        \STATE \textbf{Requirements:} max diffusion step $K$, sequence length $L$, env time limit $t_{max}$, historical condition length $T_{HC}$, state dimension $d_s$, action dimension $d_a$, reply buffer $D$, noise schedule $\alpha_{0:K}$ and $\beta_{0:K}$
        \STATE \textbf{Initialization:} $\theta$, $\phi$
        \STATE // \textbf{Prepare for Training}
        \STATE Separate the state trajectories of $D$ into state segments with length $L$
        \STATE Normalize state segments to obey uniform distribution 
        \STATE Mark the first $T_{HC}$ states as historical condition $\mathcal{C}_{HC}$ 
        \STATE Mark the action reward $r_{T_{HC}}$ of state $s_{T_{HC}}$ as immediate condition $\mathcal{C}_{IC}$
        \STATE Find the min and max trajectory return $\mathcal{T}_{min}, \mathcal{T}_{max}$ from $D$
        \STATE Mark the trajectory returns normalized with $\mathcal{T}_{min}, \mathcal{T}_{max}$ as $\mathcal{C}_{PC}$ 
        \STATE Construct the temporal condition $\mathcal{C}_{TCG}$ by wrapping $\mathcal{C}_{HC}$, $\mathcal{C}_{IC}$, and $\mathcal{C}_{PC}$
        \STATE // \textbf{Training Stage}
        \FOR{each train iteration}
            \FOR{each train step}
                \STATE Sample $b$ state sequences $\tau_{s}^{0}=\{s_{i:i+L}\}\in\mathbb{R}^{b\times L\times d_s}$ , RTGs $\tau_{RTG}=\{\mathcal{T}\}\in\mathbb{R}^{b\times 1}$, action rewards $r_a=\{\tau_{r}[i]\}\in\mathbb{R}^{b\times 1}$, and time steps $\tau_t=\{i\}\in\mathbb{R}^{b\times 1}$ from $D$
                \STATE Sample diffusion time step $k\sim \text{Uniform}(K)$
                \STATE Obtain $\tau_{s}^{k}$ by adding noise to $\tau_{s}^{0}$
                \STATE Sample gaussian noise $\epsilon \sim \mathcal{N}(0,\bm{I}), \epsilon\in\mathbb{R}^{b\times L\times d_s}$
                \STATE Train $\epsilon_{\theta}$ with $\mathcal{C}_{TCG}=[r_a, \tau_{RTG}, \tau_{t}]$ according to Equation~\ref{train DM} 
                \STATE Train $f_{\phi}$ with $\mathbb{E}_{\tau_{s}^{0},\tau_{a}^{0}}[||\tau_{a}^{0}-f_{\phi}(\tau_{s}^{0}[:,0:2,:])||^2]$
            \ENDFOR
            \STATE Save model periodically
        \ENDFOR
        \STATE // \textbf{Prepare for Evaluation}
        \STATE Select the top-$Y$ trajectories $\{\tau_i\}_{i\in Y}$ according to the returns 
        \STATE Calculate the mean reward trajectories $\tau_r$ by average on $\{\tau_{r, i}\}_{i\in Y}$ 
        \STATE $t=0$, $\mathcal{T}=\mathcal{T}_{max}$ 
        \STATE // \textbf{Evaluation Stage}
        \FOR{each evaluation step}
            \STATE Receive state $s_t$ from the environment 
            \STATE Let $k=K$ 
            \STATE Sample $\hat{\tau}^{k}_s\in\mathbb{R}^{1\times L\times d_s}$ from normal distribution $\mathcal{N}(0, \bm{I})$ 
            \STATE // Replace the first $T_{HC}$ items of $\hat{\tau}^{k}_s$ with $s_t$ \\
            \IF{$t<T_{HC}$}
                \STATE Perform sequence padding with zero states and $s_t$
            \ELSE
                \STATE Perform sequence padding with historical states and $s_t$
            \ENDIF
            \STATE Construct $\mathcal{C}_{TCG}=[\tau_r[t], \mathcal{T}, t]$ 
            \STATE Generate sequences $\hat{\tau}$ according to Equation~\ref{eval DM}
            \STATE Observe reward $r$ from the environment 
            \STATE $t=t+1$, $\mathcal{T}=\mathcal{T}-r$
        \ENDFOR 
    \end{algorithmic}
\end{algorithm}

As shown in lines $6-12$, we first process the sequences in the replay buffer to obtain the temporal conditions.
Then during the training stage (lines $14-23$), we use the diffusion model to model the joint distribution between the sequences $\tau_s$ and the temporal conditions $\mathcal{C}_{TCG}$.
After getting the well-trained diffusion model, we select the top-$Y$ trajectories and extract $\mathcal{C}_{IC}$ and $\mathcal{C}_{PC}$ (lines $26-27$).
We construct $\mathcal{C}_{HC}$ during the evaluation.
Finally, during the evaluation stage (lines $30-44$), we leverage the temporally-composable condition $\mathcal{C}_{TCG}=[\tau_r[t],\mathcal{T},t]$ to guide the sequence generation.

\section{Experimental Details}\label{Experimental Details}

\subsection{Computational Resource Description}
Experiments are carried out on NVIDIA GeForce RTX 3090 GPUs and NVIDIA A10 GPUs.
Besides, the CPU type is Intel(R) Xeon(R) Gold 6230 CPU @ 2.10GHz.
Each run of the experiments spanned about 24-48 hours, depending on the hyperparameters setting and the complexity of the environments.

\subsection{Hyperparameters}\label{Hyperparameters}

\begin{table}[ht!]
\centering
\caption{The hyperparameters of \ourmodel{}.}
\label{our method hyper}
\begin{tabular}{lc}
\toprule
Hyperparameter & Value\\
\midrule

historical sequence length $T_{HC}$ & 5 \\
max diffusion step $K$ & 200 \\
condition guidance $\omega$ & 1.2 \\
sequence length $L$ & 100 \\
network backbone & U-net~\cite{ronneberger2015u} \\
max return offset & 0/300/500 \\
loss function & MSE \\
learning rate & $2\cdot10^{-4}$ \\
batch size & 32 \\
optimizer & Adam~\cite{kingma2014adam} \\
top-$Y$ & 1/3/5/7/9 \\
$\gamma$ & 0.99 \\
\bottomrule
\end{tabular}
\vspace{-0.4cm}
\end{table}

\section{Detailed Environment Description}\label{Detail Environment Description}

\subsection{Temporal Condition Scenarios}\label{Temporal Condition Scenarios}
In order to show the temporal dependencies in sequential modeling and investigate whether the previous methods with prospective discounted returns, such as DD, and whether \ourmodel{} can generate satisfactory sequences, we design three types of RL environments, which are shown in Figure~\ref{temporal condition env results}.

In the first scenario (\textbf{historical condition env} in Figure~\ref{temporal condition env results} (a)), we define that the collected state sequences are generated according to three different historical state sequences.
We further avoid the diffusion model distinguishing state sequences only from the current state by letting these three types of historical sequences converge to the same junction state (blue circle). 
The green triangle trajectories can only be unlocked when the historical state sequence meets the condition of state incidence at the $135^{\circ}$ angle to the junction state.

We fix the historical sequences as the same circumstance but modify the action reward under the junction state in the second scenario (\textbf{immediate condition env}) in Figure~\ref{temporal condition env results} (b).
We set the low-reward trajectory samples and high-reward trajectory samples on the action reward of the junction state.
Besides, we hold the total returns of trajectories the same for these two types of trajectory samples.
As described in Section~\ref{our method}, we will obtain the generative sequences from the diffusion model, but we will not use the entire sequence for decision-making. 
Instead, we will make decisions based on $s_t$ and $s_{t+1}$.
Given the above settings, we can evaluate whether the diffusion model can focus on immediate behaviors that are most related to the current interaction step with the environment.

In the \textbf{prospective condition env} (Figure~\ref{temporal condition env results} (c)), we first separate the trajectories into two parts of equal length.
Next, we set the reward of the front part of the upper trajectories (green triangle) as 0 and the reward of the latter part as 1 for each action.
In contrast, the below trajectories (red square) have the opposite of the reward setting, i.e., 1 for the front part and 0 for the latter part.
The setting of same returns of trajectories and different action reward distributions in trajectories makes it suitable for us to test the difference between the return-guided diffusion model and the RTG-guided diffusion model.

\begin{table*}[t!]
\centering
\small
\caption{Offline RL algorithm comparison on Maze2D dataset and Hand Manipulation dataset.}
\label{table comparison of maze2d}
\resizebox{0.9\textwidth}{!}{
\begin{tabular}{l | l | l r r r}
\toprule
\specialrule{0em}{1.5pt}{1.5pt}
\toprule
model & &  & BC & DD & TCD\\
\midrule[1pt]
\multirow{6}{*}{Maze2D} & \multirow{3}{*}{Maze2D sparse} & umaze & -5.40\tiny{$\pm$13.59} & 17.16\tiny{$\pm$38.96} & 39.99\tiny{$\pm$39.61} \\
                               &  & medium & 12.35\tiny{$\pm$17.57} & -3.10\tiny{$\pm$6.99} & 28.18\tiny{$\pm$5.28} \\
                               &  & large & 3.83\tiny{$\pm$12.35} & -14.19\tiny{$\pm$6.25} & 7.68\tiny{$\pm$14.64} \\
\cline{2-6}
\rule{0pt}{2.5ex} &  \multirow{3}{*}{Maze2D dense} & umaze & -14.56\tiny{$\pm$11.42} & 83.23\tiny{$\pm$37.63} & 29.77\tiny{$\pm$21.17} \\
                               &  & medium & 16.31\tiny{$\pm$27.76} & 78.17\tiny{$\pm$83.91} & 41.44\tiny{$\pm$9.45} \\
                               &  & large & 17.09\tiny{$\pm$34.82} & 22.97\tiny{$\pm$40.06} & 75.51\tiny{$\pm$16.26} \\
\midrule[1pt]
\multicolumn{2}{c}{score}      &  & 4.49 & 30.71 & \textbf{37.10} \\
\midrule[1pt]
\multirow{6}{*}{Hand Manipulation} & \multirow{3}{*}{Pen}           & human & 7.45\tiny{$\pm$13.59} & 7.45\tiny{$\pm$13.59} & 49.88\tiny{$\pm$63.92} \\
                               &  & expert & 69.67\tiny{$\pm$61.85} & 8.99\tiny{$\pm$29.77} & 35.60\tiny{$\pm$60.74} \\
                               &  & cloned & 6.63\tiny{$\pm$31.32} & 55.12\tiny{$\pm$66.20} & 73.30\tiny{$\pm$66.48} \\
\cline{2-6}
\rule{0pt}{2.5ex} &  \multirow{3}{*}{Relocate}      & human & 0.06\tiny{$\pm$0.02} & 0.07\tiny{$\pm$0.23} & 0.35\tiny{$\pm$1.20} \\
                               &  & expert & 57.12\tiny{$\pm$46.42} & 80.31\tiny{$\pm$41.05} & 59.64\tiny{$\pm$38.35} \\
                               &  & cloned & 0.05\tiny{$\pm$0.04} & 0.07\tiny{$\pm$0.15} & 0.15\tiny{$\pm$0.34} \\
\midrule[1pt]
\multicolumn{2}{c}{score}      &  & 23.46 & 30.47 & \textbf{36.49} \\
\midrule[1pt]
\multicolumn{2}{c}{total mean score} & & 13.98 & 30.59 & \textbf{36.80} \\
\bottomrule
\specialrule{0em}{1.5pt}{1.5pt}
\bottomrule
\end{tabular}}
\end{table*}

\subsection{Gym-MuJoCo}
Apart from the environments that we introduce in Section~\ref{Evaluation with baseliens}, we also evaluate our method in other scenarios, including -random (\textbf{-r}), -expert (\textbf{-e}), and -full-replay (\textbf{-fr}).
The version of all the Gym-MuJoCo environments is \textbf{-v2}. 

\subsection{Maze2D}\label{appendix Maze2D description}
The Maze2D is a 2D navigation task where the agent needs to reach the specific location.
The goal is finding the shortest path to the evaluation location by training on a previously-collected dataset.
There are three difficulty settings, -umaze (\textbf{-u}), -medium (\textbf{-m}), and -large (\textbf{-l}), about Maze2D according to the size of the maze layouts.
The trajectories are constructed with waypoint sequences that are generated by a PD controller~\cite{fu2020d4rl}.
Apart from the above-introduced difficulty settings, there are also two reward settings: sparse reward setting and dense reward setting.
Thus we can obtain six scenarios by permutation and combination: Maze2D-sparse-u, Maze2D-sparse-m, Maze2D-sparse-l, Maze2D-dense-u, Maze2D-dense-m, and Maze2D-dense-l. 
The version of Maze2D is \textbf{-v1}.

\subsection{Hand Manipulation}
Hand Manipulation (i.e., Adroit) contains several sparse-reward, high-dimensional robotic manipulation tasks where the datasets are collected under three types of situations (-human (\textbf{-h}), -expert (\textbf{-e}), and -cloned (\textbf{-c}))~\cite{rajeswaran2017learning}.
For example, Relocate scenario requires the agent to pick up a ball and move it to a specific location, and Pen is a scenario where the agent needs to get rewards by twirling a pen.
The datasets with -h difficulty contain a small number of trajectories, while the datasets with -e and -c include abundant trajectories for training.
Compared with Maze2D and Gym-MuJoCo, Hand Manipulation possesses a higher state dimension, harder exploration, more real expert demonstration, and more sparse reward feedback.
The version of Hand Manipulation is \textbf{-v1}.

\begin{table}[t!]
\centering
\small
\caption{Additional comparison on Gym-MuJoCo dataset.}
\label{Additional MuJoCo comparison}
\resizebox{\textwidth}{!}{
\begin{tabular}{l | r r r | r r r | r r r | r}
\toprule
\specialrule{0em}{1.5pt}{1.5pt}
\toprule
Dataset & \multicolumn{3}{c|}{Random} & \multicolumn{3}{c|}{Expert} & \multicolumn{3}{c|}{Full-Replay} & \multirow{2}{*}{score}\\
\cline{1-10}
\rule{0pt}{2.5ex} Env & HalfCheetah & Hopper & Walker2d & HalfCheetah & Hopper & Walker2d & HalfCheetah & Hopper & Walker2d\\
\midrule[1pt]
DD & 1.96\tiny{$\pm$0.11} & 5.81\tiny{$\pm$1.25} & 12.84\tiny{$\pm$5.66} & 17.40\tiny{$\pm$24.55} & 110.87\tiny{$\pm$2.81} & 103.67\tiny{$\pm$17.73} & 45.64\tiny{$\pm$11.35} & 99.60\tiny{$\pm$22.30} & 82.06\tiny{$\pm$5.52} & 53.32 \\
\ourmodel{} & 3.98\tiny{$\pm$0.61} & 2.14\tiny{$\pm$0.82} & 6.50\tiny{$\pm$1.78} & 36.38\tiny{$\pm$28.70} & 112.65\tiny{$\pm$1.06} & 108.11\tiny{$\pm$0.94} & 25.65\tiny{$\pm$14.66} & 106.27\tiny{$\pm$0.82} & 96.16\tiny{$\pm$1.27} & \textbf{55.32} \\
\bottomrule
\specialrule{0em}{1.5pt}{1.5pt}
\bottomrule
\end{tabular}}
\end{table}

\begin{table}[t!]
\centering
\small
\caption{Ablation study on temporal conditions under general comparison.}
\label{table ablation results under general}
\resizebox{\textwidth}{!}{
\begin{tabular}{l | r r r | r r r | r r r | r}
\toprule
\specialrule{0em}{1.5pt}{1.5pt}
\toprule
Dataset & \multicolumn{3}{c|}{Med-Expert} & \multicolumn{3}{c|}{Medium} & \multicolumn{3}{c|}{Med-Replay} & \multirow{2}{*}{score}\\
\cline{1-10}
\rule{0pt}{2.5ex} Env & HalfCheetah & Hopper & Walker2d & HalfCheetah & Hopper & Walker2d & HalfCheetah & Hopper & Walker2d\\
\midrule[1pt]
\PCDM{} & 42.68\tiny{$\pm$24.25} & 92.98\tiny{$\pm$20.20} & 79.22\tiny{$\pm$3.08} & 41.81\tiny{$\pm$5.58} & 71.24\tiny{$\pm$18.78} & 79.22\tiny{$\pm$3.08} & 32.76\tiny{$\pm$10.07} & 91.01\tiny{$\pm$8.92} & 77.97\tiny{$\pm$10.03} & 66.45 \\
\ICDM{} & 41.81\tiny{$\pm$29.01} & 109.79\tiny{$\pm$10.28} & 108.26\tiny{$\pm$0.46} & 43.61\tiny{$\pm$5.46} & 62.31\tiny{$\pm$18.39} & 68.08\tiny{$\pm$20.13} & 38.97\tiny{$\pm$4.89} & 92.69\tiny{$\pm$9.54} & 71.49\tiny{$\pm$19.91}  & 72.33 \\ 
\HCDM{} & 76.92\tiny{$\pm$25.81} & 111.15\tiny{$\pm$5.08} & 108.75\tiny{$\pm$0.31} & 44.63\tiny{$\pm$0.63} & 71.48\tiny{$\pm$16.27} & 77.92\tiny{$\pm$7.96} & 34.24\tiny{$\pm$8.85} & 94.15\tiny{$\pm$6.49} & 75.33\tiny{$\pm$14.99} & 77.17 \\ 
\ourmodel{} & 74.30\tiny{$\pm$29.52} & 110.94\tiny{$\pm$8.99} & 111.31\tiny{$\pm$0.73} & 46.73\tiny{$\pm$1.93} & 74.48\tiny{$\pm$19.95} & 73.61\tiny{$\pm$13.26} & 35.29\tiny{$\pm$10.60} & 93.32\tiny{$\pm$8.37} & 78.83\tiny{$\pm$13.01} & \textbf{77.65} \\
\bottomrule
\specialrule{0em}{1.5pt}{1.5pt}
\bottomrule
\end{tabular}}
\vspace{-0.4cm}
\end{table}

\section{Additional Experiment Results}\label{additional experiments}

\subsection{Discussion of Temporal Conditions Scenarios}\label{Discussion of Temporal Conditions Scenarios}
The results of Temporal Conditions Scenarios are shown in the below part of Figure~\ref{temporal condition env results}. (Refer to Section~\ref{Temporal Condition Scenarios} for environmental description.)
We select the DD with prospective returns as a comparison.
For the \textbf{historical condition env} (Figure~\ref{temporal condition env results} (d)), the generation process of DD needs the current state as the condition.
When we give junction state as the condition, DD can not generate the corresponding sequences because it does not consider the temporal dependencies between history and immediate state.
Our method (\ourmodel{}) can produce the adequate trajectories that meet the historical sequence condition because of the guidance of $\mathcal{C}_{HC}$.
In the \textbf{immediate condition env} (Figure~\ref{temporal condition env results} (e)), we can see that \ourmodel{} can distinguish different action reward trajectories under the same trajectory returns with immediate condition $\mathcal{C}_{IC}$ guidance.
Thus it is obvious that \ourmodel{} can generate the action with high rewards during evaluation.   
Although DD can also generate the sequence with high action reward, the single mode of generative samples reveals poor discernibility and hinders DD from applications such as diverse trajectories generation.
Finally, in the \textbf{prospective condition env} (Figure~\ref{temporal condition env results} (f)), we can also see that the RTG with the remaining available time step can help the diffusion model recognize different trajectories.
However, the DD can not recover all types of trajectories when the prospective returns are the same.

\subsection{Additional Exploration about Temporal Conditions}\label{Additional Exploration about Temporal Conditions}

\noindent\textbf{\PCDM{}.}
Compared with previous diffusion-based methods that estimate the action value function Q or state value function V, we can obtain the RTG instruction easily from the experiences rather than suffering from inaccurate estimation on OOD samples.
Besides, RTG relates the remaining available time step with the historical best performance in one episode, which can not be reflected by the Q or V function.
Statistically, we first calculate the episode returns from the replay buffer and then obtain the maximal return value $\mathcal{T}_{max}$ and minimum return value $\mathcal{T}_{min}$.
After that, the RTG information from state $s_t$ is normalized by $\mathcal{J}_t=\frac{\mathcal{J}_t-\mathcal{T}_{min}}{\mathcal{T}_{max}-\mathcal{T}_{min}}$.
Finally, we use $[\mathcal{J}_t, t]$ as the condition during training, i.e., $\mathcal{C}_{PC}=[\mathcal{J}_t, t]$ when $PC=RTG$.

\noindent\textbf{\HCDM{}.}
Based on the U-net backbone, we add the historical information by replacing the incipient generative state segments $\{\hat{s}_t, ..., \hat{s}_{t+T_{HC}-1}\}$ of sequences with $\mathcal{H}=\{s_t, ..., s_{t+T_{HC}-1}\}$ at each generative step.
Previous methods, like DD, generate state sequence conditioning only on current state $s_t$, which may omit important information that appears in history, especially for environments with partial observability.
Even for environments without partial observability, the historical conditions can be regarded as an argumentation method, and the experiments show they can still provide improvements.
We also use RTG and remaining time steps as additional information during training. Then the historical condition is defined as $\mathcal{C}_{HC}=[\mathcal{H}, \mathcal{J}_t, t]$.

\noindent\textbf{\ICDM{}.}
During the inference stage, though we will generate a state sequence $\{\hat{s}_t\}_{t:t+T-1}$ under current state $s_t$, only state $\hat{s}_{t+1}$ is used to produce action $a_t$, which inspires us to focus on the generative quality of the current action.
Thus we adopt the first action reward $r_t$ that associates $s_t$ and $a_t$ to instruct generation, where $r_t$ is calculated from top-$Y$ trajectories in the replay buffer.
Mathematically, $\mathcal{C}_{IC}=[r_t, \mathcal{J}_t, t]$ is used for training, and $\tau_{r}=\mathbb{E}_{\tau\sim D_{\text{top-K}}}[\tau]$ is used for evaluation, where $r_t$ and $\tau_r$ are normalized according to maximal and minimum single-step reward.

\begin{figure*}[t!]
 \begin{center}
 \includegraphics[angle=0,width=0.99\textwidth]{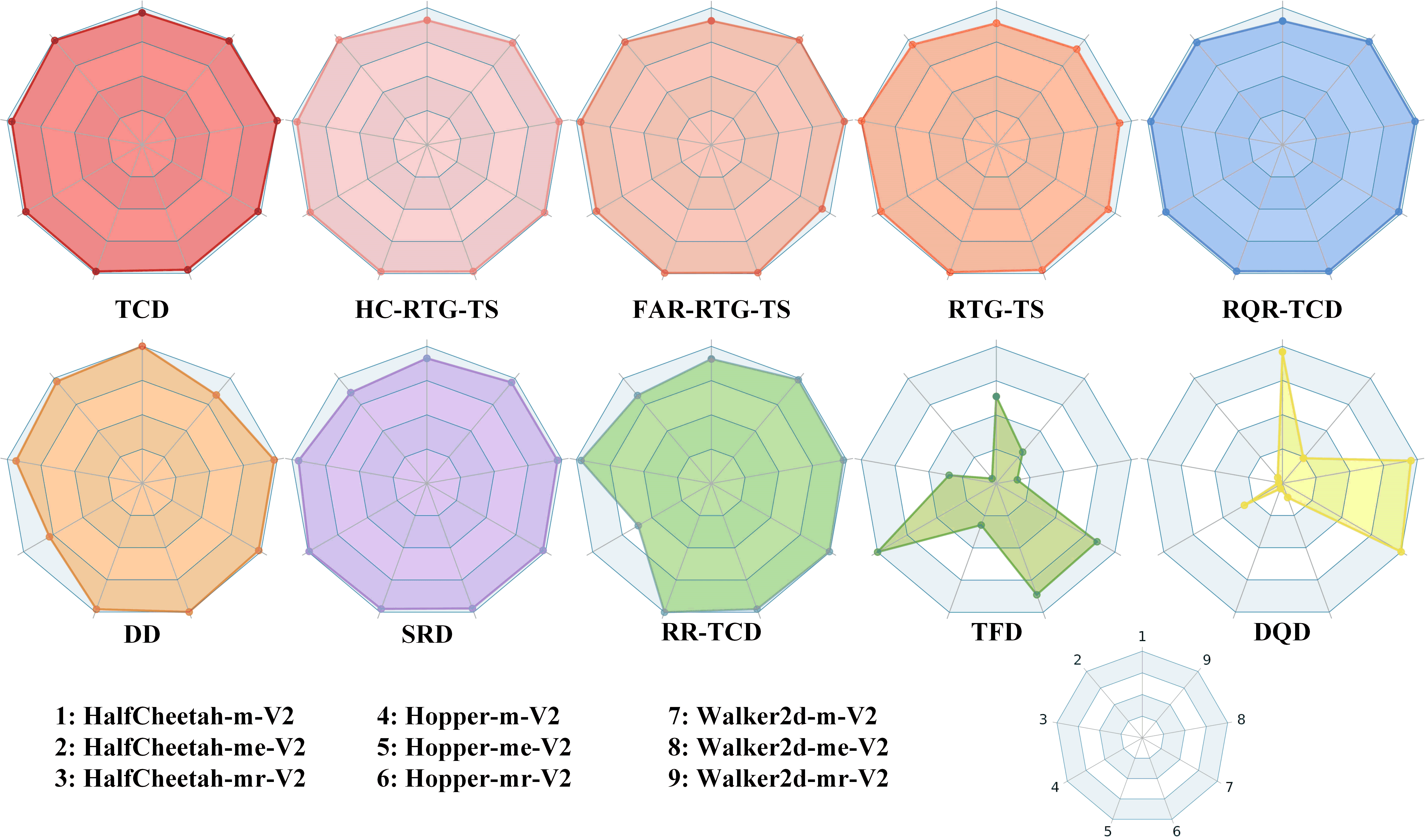}
 \caption{Evaluation of various temporal methods on D4RL Gym-MuJoCo dataset under successful comparison. The area of the polygon represents holistic performance. Each vertex of the polygon denotes the normalized performance across all methods.}
 \label{radar comparison on all temporal conditions}
 \end{center}
 \vspace{-0.3cm}
 \end{figure*}

\begin{table*}[t!]
\centering
\small
\caption{Successful comparison of the effect of various temporal conditions on diffusion model.}
\label{table comparison on all temporal conditions under general comparison}
\resizebox{\textwidth}{!}{
\begin{tabular}{l | r r r | r r r | r r r | r}
\toprule
\specialrule{0em}{1.5pt}{1.5pt}
\toprule
Dataset & \multicolumn{3}{c|}{Med-Expert} & \multicolumn{3}{c|}{Medium} & \multicolumn{3}{c|}{Med-Replay} & \multirow{2}{*}{score}\\
\cline{1-10}
\rule{0pt}{2.5ex} Env & HalfCheetah & Hopper & Walker2d & HalfCheetah & Hopper & Walker2d & HalfCheetah & Hopper & Walker2d\\
\midrule[1pt]
\DistributionalQDiffuser{} & 5.05\tiny{$\pm$1.92} & 4.79\tiny{$\pm$0.33} & 106.03\tiny{$\pm$1.39} & 47.09\tiny{$\pm$2.00} & 32.55\tiny{$\pm$0.04} & 84.07\tiny{$\pm$0.63} & 1.16\tiny{$\pm$0.91} & 11.06\tiny{$\pm$0.19} & 21.35\tiny{$\pm$0.09} & 34.79\\
\TFBB{} & 4.47\tiny{$\pm$2.28} & 36.78\tiny{$\pm$29.13} & 17.31\tiny{$\pm$5.17} & 31.23\tiny{$\pm$8.36} & 101.42\tiny{$\pm$0.85} & 71.36\tiny{$\pm$2.85} & 14.68\tiny{$\pm$5.06} & 86.37\tiny{$\pm$2.01} & 26.69\tiny{$\pm$11.49} & 43.37 \\
\RRDM{} & 78.31\tiny{$\pm$28.93} & 114.23\tiny{$\pm$1.09} & 108.89\tiny{$\pm$0.23} & 44.60\tiny{$\pm$0.93} & 62.46\tiny{$\pm$13.15} & 83.60\tiny{$\pm$1.44} & 40.55\tiny{$\pm$1.75} & 97.59\tiny{$\pm$1.27} & 88.05\tiny{$\pm$2.24} & 79.81\\
\PCDM{} & 89.11\tiny{$\pm$4.47} & 113.13\tiny{$\pm$1.55} & 101.60\tiny{$\pm$9.90} & 43.58\tiny{$\pm$1.26} & 98.66\tiny{$\pm$1.03} & 79.23\tiny{$\pm$3.09} & 42.01\tiny{$\pm$1.23} & 97.27\tiny{$\pm$0.00} & 81.26\tiny{$\pm$4.30} & 82.87\\
\StateRewardDiffuser{} & 80.79\tiny{$\pm$25.59} & 111.40\tiny{$\pm$0.62} & 107.72\tiny{$\pm$0.72} & 44.81\tiny{$\pm$0.77} & 100.58\tiny{$\pm$0.02} & 82.29\tiny{$\pm$2.11} & 39.95\tiny{$\pm$1.33} & 97.14\tiny{$\pm$1.52} & 85.76\tiny{$\pm$2.74} & 83.38 \\
\ICDM{} & 91.56\tiny{$\pm$2.80} & 113.63\tiny{$\pm$1.98} & 109.62\tiny{$\pm$0.66} & 44.48\tiny{$\pm$0.72} & 98.05\tiny{$\pm$2.00} & 78.55\tiny{$\pm$3.65} & 40.61\tiny{$\pm$1.58} & 99.21\tiny{$\pm$1.59} & 89.18\tiny{$\pm$2.65} & 84.98\\
\RQRDM{} & 90.97\tiny{$\pm$2.88} & 112.22\tiny{$\pm$1.23} & 109.24\tiny{$\pm$0.20} & 44.34\tiny{$\pm$1.13} & 99.71\tiny{$\pm$0.97} & 82.24\tiny{$\pm$2.32} & 41.01\tiny{$\pm$2.28} & 98.30\tiny{$\pm$1.40} & 87.54\tiny{$\pm$3.07} & 85.06 \\
\HCDM{} & 93.44\tiny{$\pm$1.62} & 112.51\tiny{$\pm$0.46} & 108.75\tiny{$\pm$0.31} & 44.63\tiny{$\pm$0.63} & 99.91\tiny{$\pm$0.35} & 83.16\tiny{$\pm$1.70} & 40.49\tiny{$\pm$1.42} & 98.34\tiny{$\pm$1.61} & 86.57\tiny{$\pm$2.11} & 85.31\\
\ourmodel{} & 92.67\tiny{$\pm$3.37} & 112.60\tiny{$\pm$1.03} & 111.31\tiny{$\pm$0.73} & 47.20\tiny{$\pm$0.74} & 99.37\tiny{$\pm$0.60} & 82.06\tiny{$\pm$1.83} & 40.57\tiny{$\pm$1.39} & 97.20\tiny{$\pm$2.39} & 88.04\tiny{$\pm$1.92} & \textbf{85.67}\\
\bottomrule
\specialrule{0em}{1.5pt}{1.5pt}
\bottomrule
\end{tabular}}
\vspace{-0.4cm}
\end{table*}

\noindent\textbf{Diffuser with Transformer-Backbone (TFD).}~~~~
For the implicit line about historical condition, we try to replace the U-net backbone with the transformer backbone to focus on longer history information~\cite{touvron2023llama}.
We call this method diffuser with transformer-backbone (TFD), which has lower memory consumption and shorter run time because it does not use the convolution operation.

\noindent\textbf{Diffuser with State-Reward Sequence Modeling (SRD).}~~~~
For the implicitly immediate or prospective condition, we propose modeling the distribution of state sequences and the joint distribution of state sequences and the reward sequences. 
For the state $s_t$ of current time step, we use SRD to generate several candidate sequences $\{s_t, \hat{r}_t, \hat{s}_{t+1}, \hat{r}_{t+1}, ..., \hat{r}_{t+L-1}\}_{n=1}^{N}$ and select the sequence with max reward $\hat{r}_t^{n}$ (implicitly immediate condition) or max returns $\sum_{i=1}^{t+L-1}\hat{r}_i^n$ (implicitly prospective condition). 

\noindent\textbf{Diffuser with Distributional Q Estimation (DQD).}~~~~
We also conduct other explorations about explicitly-temporal conditions.
In order to obtain better approximation by preserving the multimodality of action value function, we explore fitting the Q function on offline datasets with distributional RL techniques~\cite{bellemare2017distributional}.
Specifically, we separate the discounted returns into 201 bins and use two networks to predict the bin values and the bin distribution, respectively.
Then the expected bin values weighted by the probabilities of bin selection are adopted as prospective returns to guide the diffusion model. 

\noindent\textbf{\ourmodel{} with Reward Estimation (RR-\ourmodel{} and RQR-\ourmodel{}).}~~~~
Directly utilizing statistical rewards from the replay buffer can not reflect the reward situation of all collected states. Thus we hope to improve the prediction accuracy by introducing reward estimation.
In practice, directly estimating reward values conditioning on states requires learning a mapping from states to rewards, which we call that \ourmodel{} with linear reward regression (RR-\ourmodel{}).
Another method, \ourmodel{} with reward quantile regression (RQR-\ourmodel{}), gives us a chance to roughly identify the radical and conservative actions through the reward distribution on each state.

In Appendix~\ref{More Discussion about Temporal Conditions}, we review previous studies and discuss the advantages and disadvantages of the above methods, \ourmodel{}, and other baselines.

 \begin{figure*}[t!]
 \begin{center}
 \includegraphics[angle=0,width=0.99\textwidth]{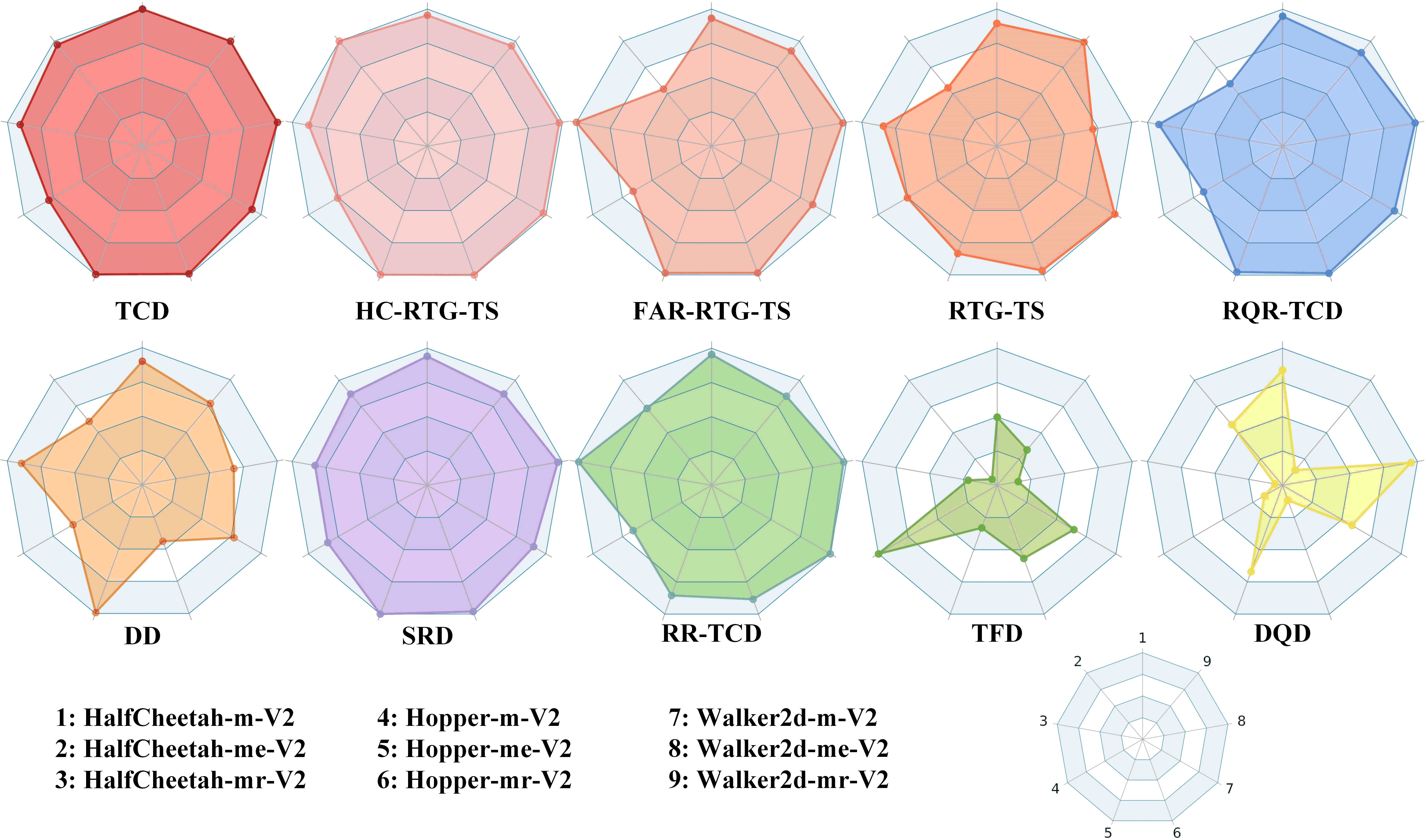}
 \caption{Evaluation of various temporal methods on D4RL Gym-MuJoCo dataset under general comparison. The area of the polygon represents holistic performance. Each vertex of the polygon denotes the normalized performance across all methods.}
 \label{radar comparison on all temporal conditions under general comparison}
 \end{center}
 \end{figure*}

\begin{table*}[t!]
\centering
\small
\caption{General comparison of the effect of various temporal conditions on diffusion model.}
\label{table comparison on all temporal conditions under successful comparison}
\resizebox{\textwidth}{!}{
\begin{tabular}{l | r r r | r r r | r r r | r}
\toprule
\specialrule{0em}{1.5pt}{1.5pt}
\toprule
Dataset & \multicolumn{3}{c|}{Med-Expert} & \multicolumn{3}{c|}{Medium} & \multicolumn{3}{c|}{Med-Replay} & \multirow{2}{*}{score}\\
\cline{1-10}
\rule{0pt}{2.5ex} Env & HalfCheetah & Hopper & Walker2d & HalfCheetah & Hopper & Walker2d & HalfCheetah & Hopper & Walker2d\\
\midrule[1pt]
\DistributionalQDiffuser{} & 4.43\tiny{$\pm$2.49} & 4.79\tiny{$\pm$0.33} & 106.03\tiny{$\pm$1.39} & 39.28\tiny{$\pm$15.12} & 14.44\tiny{$\pm$8.13} & 46.68\tiny{$\pm$18.69} & 2.22\tiny{$\pm$2.45} & 11.06\tiny{$\pm$0.19} & 11.51\tiny{$\pm$6.50} & 26.77\\
\TFBB{} & 4.47\tiny{$\pm$2.28} & 36.78\tiny{$\pm$29.13} & 17.31\tiny{$\pm$5.17} & 23.27\tiny{$\pm$9.48} & 94.47\tiny{$\pm$12.93} & 51.55\tiny{$\pm$27.52} & 8.38\tiny{$\pm$6.24} & 53.43\tiny{$\pm$30.23} & 26.69\tiny{$\pm$11.49} & 35.62\\
\PCDM{} & 42.68\tiny{$\pm$24.25} & 92.98\tiny{$\pm$20.20} & 79.22\tiny{$\pm$3.08} & 41.81\tiny{$\pm$5.58} & 71.24\tiny{$\pm$18.78} & 79.22\tiny{$\pm$3.08} & 32.76\tiny{$\pm$10.07} & 91.01\tiny{$\pm$8.92} & 77.97\tiny{$\pm$10.03} & 66.45 \\
\RRDM{} & 56.35\tiny{$\pm$25.84} & 95.33\tiny{$\pm$25.74} & 108.89\tiny{$\pm$0.23} & 44.60\tiny{$\pm$0.93} & 62.46\tiny{$\pm$13.15} & 79.74\tiny{$\pm$9.77} & 38.38\tiny{$\pm$6.52} & 83.36\tiny{$\pm$17.77} & 66.96\tiny{$\pm$25.99} & 70.73\\
\RQRDM{} & 45.88\tiny{$\pm$24.32} & 108.87\tiny{$\pm$12.84} & 109.24\tiny{$\pm$0.20} & 44.34\tiny{$\pm$1.13} & 62.93\tiny{$\pm$16.79} & 75.15\tiny{$\pm$14.39} & 35.73\tiny{$\pm$10.20} & 92.84\tiny{$\pm$11.92} & 70.35\tiny{$\pm$25.78} & 71.70\\
\ICDM{} & 41.81\tiny{$\pm$29.01} & 109.79\tiny{$\pm$10.28} & 108.26\tiny{$\pm$0.46} & 43.61\tiny{$\pm$5.46} & 62.31\tiny{$\pm$18.39} & 68.08\tiny{$\pm$20.13} & 38.97\tiny{$\pm$4.89} & 92.69\tiny{$\pm$9.54} & 71.49\tiny{$\pm$19.91}  & 72.33 \\ 
\StateRewardDiffuser{} & 66.85\tiny{$\pm$31.52} & 111.40\tiny{$\pm$0.62} & 107.72\tiny{$\pm$0.72} & 43.96\tiny{$\pm$4.64} & 79.17\tiny{$\pm$18.36} & 71.50\tiny{$\pm$16.40} & 32.38\tiny{$\pm$10.09} & 92.33\tiny{$\pm$10.20} & 68.61\tiny{$\pm$17.20} & 74.88 \\
\HCDM{} & 76.92\tiny{$\pm$25.81} & 111.15\tiny{$\pm$5.08} & 108.75\tiny{$\pm$0.31} & 44.63\tiny{$\pm$0.63} & 71.48\tiny{$\pm$16.27} & 77.92\tiny{$\pm$7.96} & 34.24\tiny{$\pm$8.85} & 94.15\tiny{$\pm$6.49} & 75.33\tiny{$\pm$14.99} & 77.17 \\
\ourmodel{} & 74.30\tiny{$\pm$29.52} & 110.94\tiny{$\pm$8.99} & 111.31\tiny{$\pm$0.73} & 46.73\tiny{$\pm$1.93} & 74.48\tiny{$\pm$19.95} & 73.61\tiny{$\pm$13.26} & 35.29\tiny{$\pm$10.60} & 93.32\tiny{$\pm$8.37} & 78.83\tiny{$\pm$13.01} & \textbf{77.65} \\
\bottomrule
\specialrule{0em}{1.5pt}{1.5pt}
\bottomrule
\end{tabular}}
\vspace{-0.4cm}
\end{table*}

\subsection{Additional Evaluation on Various Scenarios}\label{Evaluation on Various Scenarios}

In this section, we report the additional evaluation on many environments, such as Pen-\{h, e, c\}-v1, Relocate-\{h, e, c\}-v1, HalfCheetah-\{r, e, fr\}-v2, Hopper-\{r, e, fr\}-v2, and Walker2d-\{r, e, fr\}-v2.
From the results that are shown in Table~\ref{table comparison of maze2d} and Table~\ref{Additional MuJoCo comparison}, we can see that our method (\ourmodel{}) achieves a 20\% overall performance gain compared with DD and reaches the best mean performance in Pen and Relocate environments.
In the HalfCheetah-\{r, e, fr\}-v2, Hopper-\{r, e, fr\}-v2, and Walker2d-\{r, e, fr\}-v2 environments, the performance of our method surpasses DD in 5 of 6 datasets with countable modality (i.e., expert and full-replay) and 6 of 9 on all datasets.
The reason for poor performance in \{HalfCheetah, Hopper, Walker2d\} random datasets is that samples with random interaction do not possess primary modality on data distribution, thus leading to random update direction when we use the diffusion model to capture the data distribution.
Finally, the diffusion model can not learn to generate good behaviors that align with certain behavior policies according to experiences.

Besides, we also realize several algorithms, such as DQD, TFD, RR-TCD, RQR-TCD, and SRD, by considering other types of temporal conditions, and the corresponding results on classical Gym-MuJoCo datasets are shown in Figure~\ref{radar comparison on all temporal conditions} and Figure~\ref{radar comparison on all temporal conditions under general comparison}.
Although distributional Q estimation can alleviate the influence of OOD actions, the results show that directly adopting the distributional Q value as the instruction on conditional diffusion models may hurt the performance, which stimulates us to find a better way to combine the distributional techniques and diffusion models.
We can lightweight the memory overhead and reduce the time consumption of training diffusion models with transformer backbone, but we should realize that the position embedding and diffusion process will introduce two different time encoding vectors, which may conflict with each other and impact the model learning.
As shown in Table~\ref{table comparison on all temporal conditions under general comparison} and Table~\ref{table comparison on all temporal conditions under successful comparison}, TFD (diffuser with transformer backbone) performs poorly in most tasks, where the only difference is the Hopper-m task.
The reason is that the time encoding vectors may associate with the data distribution and exactly make a positive effect on the Hopper-m task.
Considering the methods, RR-TCD, RQR-TCD, and SRD, that estimate the action rewards, we find that modeling the distribution of reward sequence (SRD) is better than learning a mapping from state space to reward space (RR-TCD).
Reward quantile regression (RQR-TCD) performs better than RR-TCD and SRD because quantile regression is insensitive to certain extreme reward values that refer to radical actions and conservative actions. 
Thus, more likely, we can recover the behavior policies and reach better performance conditioning on the median reward values.

\begin{figure*}[t!]
 \begin{center}
 \includegraphics[angle=0,width=0.99\textwidth]{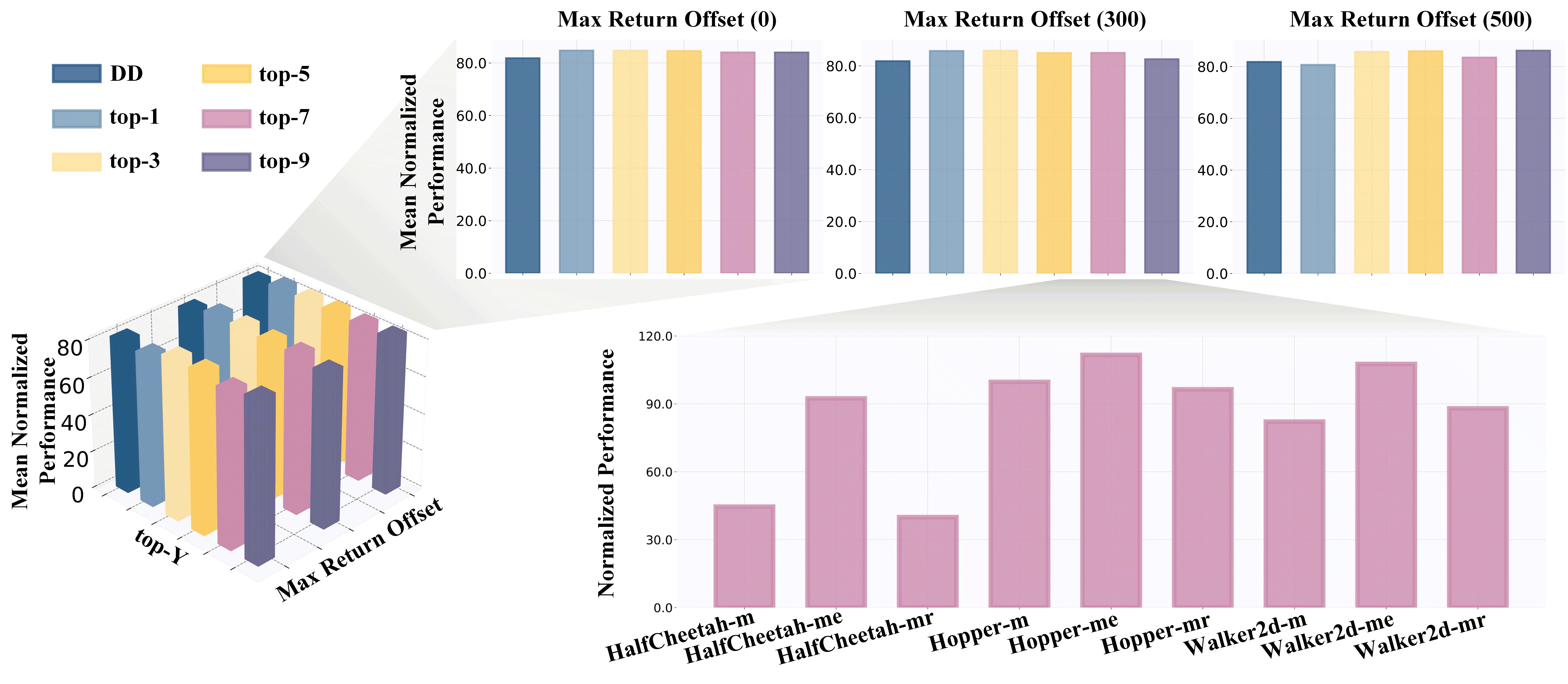}
 \caption{Parametric sensitivity about top-$Y$ and max return offset under successful comparison.}
 \label{sensitivity of top-K under successful comparison}
 \end{center}
 \end{figure*}

\begin{table}[t!]
\centering
\small
\caption{The effects of historical sequence length $L$ under successful comparison.}
\label{sensitivity of historical sequence length under successful comparison}
\resizebox{\textwidth}{!}{
\begin{tabular}{l | r r r | r r r | r r r | r}
\toprule
\specialrule{0em}{1.5pt}{1.5pt}
\toprule
Dataset & \multicolumn{3}{c|}{Med-Expert} & \multicolumn{3}{c|}{Medium} & \multicolumn{3}{c|}{Med-Replay} & \multirow{2}{*}{score}\\
\cline{1-10}
\rule{0pt}{2.5ex} Env & HalfCheetah & Hopper & Walker2d & HalfCheetah & Hopper & Walker2d & HalfCheetah & Hopper & Walker2d\\
\midrule[1pt]
\ourmodel{} ($L$5) & 92.67\tiny{$\pm$3.37} & 112.60\tiny{$\pm$1.03} & 111.31\tiny{$\pm$0.73} & 47.20\tiny{$\pm$0.74} & 99.37\tiny{$\pm$0.60} & 82.06\tiny{$\pm$1.83} & 40.57\tiny{$\pm$1.39} & 97.20\tiny{$\pm$2.39} & 88.04\tiny{$\pm$1.92} & \textbf{85.67}\\
\ourmodel{} ($L$10) & 85.74\tiny{$\pm$13.68} & 110.71\tiny{$\pm$0.50} & 109.06\tiny{$\pm$0.59} & 44.50\tiny{$\pm$0.92} & 99.60\tiny{$\pm$1.03} & 82.53\tiny{$\pm$1.32} & 39.24\tiny{$\pm$1.57} & 97.17\tiny{$\pm$1.56} & 85.06\tiny{$\pm$5.10} & 83.73\\
\bottomrule
\specialrule{0em}{1.5pt}{1.5pt}
\bottomrule
\end{tabular}}
\vspace{-0.4cm}
\end{table}

\subsection{Additional Experiment Results about Parameter Sensitivity}\label{Additional Experiments about Parameter Sensitivity}

We report the additional experiments of parameter sensitivity from two dimensions, i.e., top-$Y$ and max return offset, in Figure~\ref{sensitivity of top-K under successful comparison}, where the results show that our method performs better than DD in most settings of hyperparameters.
Besides, we also conduct experiments of parameter sensitivity on performance extrapolation and historical sequence length.

\begin{figure*}[t!]
 \begin{center}
 \includegraphics[angle=0,width=0.99\textwidth]{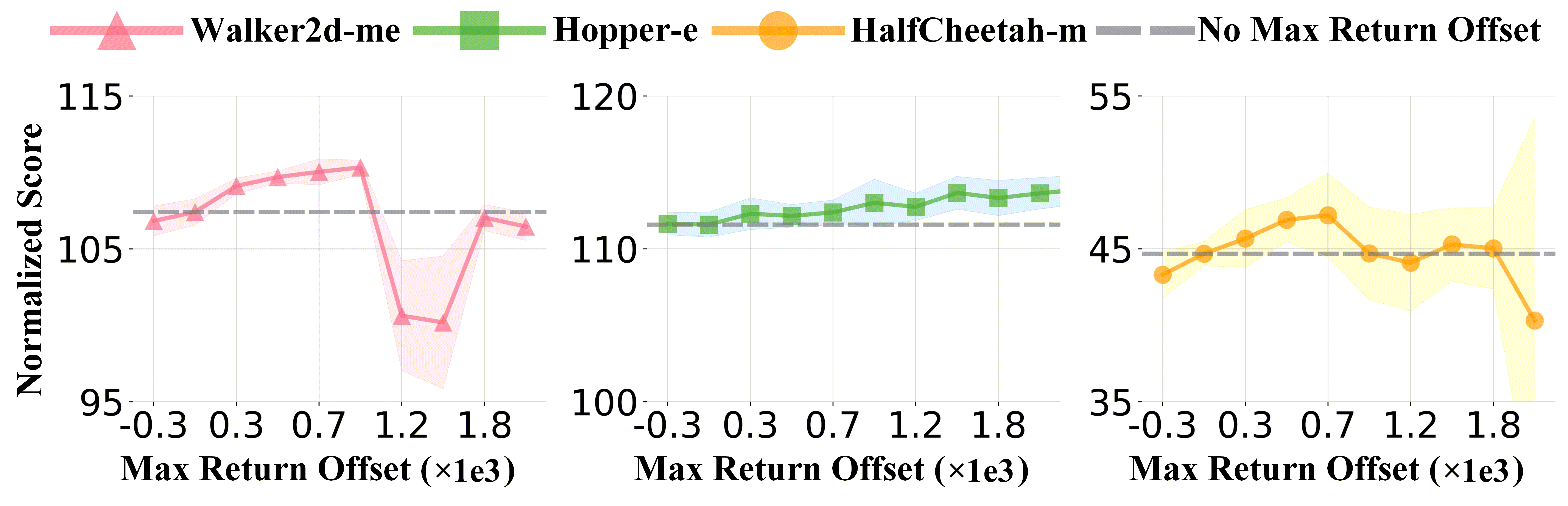}
 \caption{The performance extrapolation on max return offset.}
 \label{max return offset of max return offset}
 \end{center}
 \vspace{-0.4cm}
 \end{figure*}

\noindent\textbf{Performance Extrapolation with Max Return Offset.}~~~~
During the evaluation stage, we add max return offset to the initial RTG, where the bigger values of max return offset denote more optimism about future returns, and smaller values of max return offset indicate more pessimism about available returns.
When the condition aligns well with the training data, we can appropriately increase the condition, thereby encouraging the model to discover better decision sequences from historical experiences.
In several scenarios, We see the corresponding extrapolated phenomena that are shown in Figure~\ref{max return offset of max return offset}.
The results show that we can obtain higher performance by slightly increasing the initial RTG, which inspires us to investigate adaptive methods for selecting the max return offset.

\noindent\textbf{Historical Sequence Length $L$.}~~~~
We first probe the impacts of historical sequence length $L$, which represents how long the previous sequence is considered for generation when we adopt the U-net backbone. 
The results are shown in Table~\ref{sensitivity of historical sequence length under successful comparison}, where we can see that a relatively longer length of historical sequence can provide further improvements in Hopper-m and Walker2d-m.
While mother scenarios, the longer historical sequence ($L=10$) provides negative improvements because the historical sequence may introduce useful information and extra noise concurrently, where the useless noise makes it hard for learning control.

\section{More Discussion about Temporal Conditions}\label{More Discussion about Temporal Conditions}

\noindent\textbf{Prospectively-Conditional Sequence Generation.}
For each $\{\hat{s}_t, \hat{a}_t, \hat{r}_t\}_{t:t+T-1}$ sequence to be waiting for generation, the prospective condition information can be provided as guidance, such as expected discounted return of state value or state-action value, RTG, and target goal state.
For example, Diffuser~\cite{janner2022planning} trains a Q value function separately and adds the gradient with respect to state action pairs to guide the generation, where the stochastic sampling process is $\tau^{k-1}\sim\mathcal{N}(\mu_{\theta}+\alpha\Sigma^{k}\nabla\mathcal{J}(\mu_{\theta}), \Sigma^{k})$. $\alpha$ is the gradient scale and $\mathcal{J}=\sum_t r(s_t,a_t)$.
Limited by the restricted experiences and overestimation problem, direct Q value estimation can not provide a better approximation, while distributional RL technologies may be useful to capture the multimodal Q distribution, which leads to more stable learning~\cite{bellemare2017distributional, tian2023optimistic}.

The RTG, as another prospective condition form, indicates the future desired returns when preestablishing the initial returns, which is different from the Q value function because we obtain the Q value function through temporal difference (TD) learning while the initial returns are defined according to prior knowledge of environments.
In addition to the prospective conditions, goal state (GS) or goal feature (GF) can also be used to guide generation.
Similar to the processing way in Diffuser, where they apply $s_t$ condition by replacing the denoised sequence $\{\hat{s}_t, \hat{a}_t, \hat{r}_t\}_{t:t+T-1}$ with $\{s_t, \hat{a}_t, \hat{r}_t, \{\hat{s}_{t+1}\hat{a}_{t+1}, \hat{r}_{t+1}\}_{t+1:t+T-1}\}$, we can also substitute the final generative state $\hat{s}_{t+T-1}$ with goal state $\hat{s}_{g}$, i.e., $\{s_t, \hat{a}_t, \hat{r}_t, \{\hat{s}_{t+1}\hat{a}_{t+1}, \hat{r}_{t+1}\}_{t+1:t+T-2}, s_g, \hat{a}_{t+T-1}, \hat{r}_{t+T-1}\}$. 
Further to say, the goal state can also be embedded in latent space, so we can use $f_s$ as a general function that represents feature mapping or identity function, i.e., $s_g=f_s^{-1}(g)$ and $g=f_s(s)$.

Most previous studies adopt prospective conditions, and the general objective function is defined as
\begin{equation*}
    \mathcal{L}(\theta)=\mathbb{E}_{k\sim U(1, 2, ..., T_d), \epsilon\sim\mathcal{N}(0,\bm{I}), \tau^0\sim D}[||\epsilon-\epsilon_{\theta}(f_{\tau}(\tau^k),\mathcal{C}_{PC}, k)||_2^2],
\end{equation*}
where $\mathcal{C}_{PC}\in\{Q, RTG, GS, GF\}$ and $f_{\tau}$ is a sequence processing function, which can represent state sequence, state-action sequence, state-reward sequence, or the state-action-reward sequence.

\noindent\textbf{Historically-Conditional Sequence Generation.}
When we regard the decision-making problem as a sequence modeling problem, generating sequences based on $s_t$ is similar to long-term series forecasting, which motivates us to add historical information into sequence generation or preprocess the sequence data, such as extracting trend variables and seasonal variables firstly~\cite{wu2021autoformer, wang2023micn}.
Classical diffusion model structure utilizes the U-net backbone and one-dimensional convolution to process sequence data, so the most straightforward method to consider historical information is conditioning on preceding experiences.
From another view, the U-net backbone disregards the temporal information between consecutive transitions in a sequence, treating them as a single entity. 
In this context, if the sequence is considered as an image, the generation process with the U-net backbone can be likened to image inpainting.

In addition to the aforementioned methods that explicitly consider historical sequences, we can also implicitly take into account historical information.
Transformers employ a novel self-attention mechanism that captures long-range dependencies and global context more effectively, leading to the widespread adoption in machine translation, sentiment analysis, question-answering, and more~\cite{vaswani2017attention, devlin2018bert, radford2018improving, ouyang2022training}.
Inspired by this, we can utilize the transformer backbone to preserve longer history information rather than the U-net backbone~\cite{bao2022all, shang2021starformer}. 
When utilizing a Transformer backbone, we observe that the model is comparatively more lightweight than a U-net backbone, with reduced training time and memory overhead. 
However, it is worth noting that since the Transformer requires positional encoding for sequences, and the training process of the diffusion model necessitates the inclusion of diffusion time step information, there may be interference between these two temporal aspects.

\noindent\textbf{Immediately-Conditional Sequence Generation.}
Though we can use the diffusion model to plan a long-term sequence, only the first two states $s_{t}$ and $\hat{s}_{t+1}$ are adopted to produce actions with inverse dynamics $a_t=f_{inv}(s_{t}, \hat{s}_{t+1})$.
Consequently, the direct influencing factor in obtaining rewards from the environment is the quality of the generated states $\hat{s}_{t+1}$.
This enlightens us that we should pay more attention to the current generative state $\hat{s}_{t+1}$.

Immediate conditions are further categorized into two distinct types: those based on post-hoc filtering and those based on prior guidance.
For the post-hoc filtering method, we can use state-reward sequences and filtrate high-quality sequences on the basis of multi-candidate sequences, while most previous works choose state sequences or state-action sequences for training.
Prior guidance methods require reward sequence $\{r_t\}_{t:t+T-1}$ as guidance so as to instruct the generation process, i.e., reward sequence statistic from replay buffer is the most straightforward idea. 
Alternatively, reward regression (linear regression and quantile regression) is another choice.
Although reward regression methods can offer a degree of extrapolation capabilities, they may also be prone to overfitting, resulting in large estimation biases for OOD actions~\cite{geman1992neural, zhang2021understanding}. 
On the other hand, statistical reward methods directly utilize historical experience for guidance, which may, to some extent, constrain the model's performance~\cite{bengio2009curriculum, cobbe2019quantifying}.

\section{More Discussion about Limitations and Future Work}\label{More Discussion about Limitation and Future Work}
In terms of limitations, the mechanism of generation process makes it slower than other models, such as Transformer-based models and MLP-based models, even though we can use recent breakthroughs~\cite{nichol2021improved} to accelerate this process.
More recent studies have brought hope for efficient generation. Thus we may be able to improve the efficiency based on the current models~\cite{song2023consistency}.
Another limitation is the restricted application on static datasets such as offline RL tasks because, in these static datasets, the joint distribution of samples is fixed.
While in online learning, the update of behavior policies influences the data distribution collected from the environment.

\end{document}